# RAVEN: Frozen Random Graph Reservoirs with Physics-Informed Interaction Fingerprints for Protein–Ligand Binding Affinity Prediction

*Qingyang Zou[1], Jiaye Huang[2], Hangbo Xie[1], Jiayue Yin[1], Youyi Song[1]*, and Jinfeng Liu[1,2]**

*[1]School of Science, China Pharmaceutical University, Nanjing, 210009, China*

*[2]School of Basic Medicine and Clinical Pharmacy, China Pharmaceutical University, Nanjing, 210009, China*

*To whom correspondence should be addressed: jinfengliu@cpu.edu.cn (J.L.), youyisong@cpu.edu.cn

## Abstract

Quantitative estimation of protein–ligand binding affinity from three-dimensional complex structures is a fundamental task in structure-based computational chemistry and molecular modeling. Reliable prediction remains challenging because available structure–affinity data are limited, experimentally heterogeneous, conformation-dependent, and sensitive to dataset partitioning. RAVEN (Randomized Atomistic Views with Ensemble Neural Reservoirs) utilizes a multi-head reservoir of independently initialized and fully frozen atomistic graph encoders to generate diverse structural projections without end-to-end optimization of the graph representation. These projections are integrated with a deterministic physicochemical interaction fingerprint and processed by heterogeneous supervised readers, including neural and tree-based regressors, whose outputs are combined through validation-based nonnegative fusion. The random reservoir expands structural feature coverage across independent encoder realizations, whereas the explicit physicochemical descriptors and heterogeneous readers contribute complementary information and distinct inductive biases. Evaluation on a similarity-isolated PDBbind 2020R1 split reconstructed using GEMS similarity resources, together with the protected CASF-2016 subset, demonstrated strong predictive performance. The results indicate that frozen multi-view graph representations, explicit physicochemical statistics, and heterogeneous model fusion provide a robust and flexible framework for protein–ligand binding-affinity prediction.

## Introduction

Three-dimensional structures reveal how a ligand occupies a protein binding site, but the observed bound geometry alone does not determine the thermodynamic strength of association, which reflects contributions from intermolecular interactions, solvation, conformational reorganization, and entropy[1–3]. Structure-based binding-affinity prediction therefore seeks to translate the atomic organization of a protein–ligand complex into an experimental measure of binding strength, commonly an equilibrium constant such as $K_d$ or $K_i$, or a corresponding logarithmic or free-energy representation[4–6]. Resources such as PDBbind have made this problem accessible to data-driven modeling by pairing experimentally characterized affinities with three-dimensional complex structures, yet structurally resolved affinity data remain limited and heterogeneous measurements may combine equilibrium constants with assay-dependent $IC_{50}$ values[4–8]. The structural input is also not unique: crystallographic, redocked, and cross-docked conformations can encode different local contact patterns, while established benchmarking frameworks distinguish affinity scoring from the separate problem of identifying or ranking binding poses[9,10]. Binding-affinity prediction is therefore not a direct reading of static geometry, but a data-limited regression problem whose apparent performance can depend jointly on experimental labels, structural quality, conformational representation, and the relationship between training and evaluation data[5,6,8,11].

Structure-based machine-learning scoring functions have adopted a broad range of molecular representations, including predefined interaction descriptors, three-dimensional neural inputs, and atom-level graph models[5,6,12,13]. RF-Score showed that compact protein–ligand atom-pair contact counts could support effective nonlinear affinity prediction when coupled with Random Forest regression[14]. ECIF increased the chemical resolution of this strategy by defining atom types according to their local connectivity, demonstrating that detailed fixed descriptors remain highly informative for tree-based models[15]. Deep structure-based approaches subsequently transferred part of the representation process to trainable neural architectures, allowing intermediate molecular features to be extracted directly from three-dimensional and graph-structured inputs[5,6,12,13]. PotentialNet employed staged message passing, first updating atomic representations along covalent bonds and then incorporating distance-defined molecular neighborhoods to learn higher-level interaction features[16]. InteractionGraphNet similarly used successive graph-convolution modules to learn intramolecular and intermolecular information from three-dimensional protein–ligand complexes[17]. GraphDelta followed a hybrid formulation in which invariant Behler–Parrinello symmetry functions described the protein environment around each ligand atom before message passing was performed through the ligand graph[18]. Physics-informed models further enriched graph representations with chemically structured prediction functions; PIGNet combined learned atomic features with parameterized van der Waals, hydrogen-bond, metal–ligand, and hydrophobic interaction components[19]. Together, these developments place molecular representation at the center of affinity-model design and illustrate complementary roles for chemically resolved descriptors, learned spatial representations, adaptive graph propagation, physically motivated interaction features, and nonlinear supervised readers[5,6,17].

In end-to-end affinity models, encoder parameters are optimized under the supervised affinity objective, allowing the learned representation to reflect both useful molecular information and regularities of the training distribution[6,11,20,21]. Such flexibility can be valuable, but similarity and redundancy across training and evaluation data can also reward memorization or other predictive shortcuts, producing performance estimates that overstate generalization to less related chemical or protein space[8,11,22–24]. Modular message-passing studies found that ligand-only or protein-only representations could match or exceed explicit intermolecular interaction graphs, while simple similarity-based estimates approached the performance of learned graph models[20]. Independent analysis of PDBbind likewise found ligand-only and protein-only models to retain substantial predictive performance relative to the corresponding full-complex model, and similarity-aware splits produced more demanding estimates of model generalizability[11]. Related perturbation studies on kinase affinity benchmarks found substantially greater sensitivity to ligand encodings than to the investigated protein encodings, further illustrating that high predictive accuracy does not by itself establish balanced use of both molecular partners[6]. Explainability analyses across multiple GNN architectures have similarly identified strong contributions from ligand-associated training patterns and architecture-dependent use of intermolecular interaction edges[21]. Learned affinity representations may therefore contain both interaction-related information and dataset-specific regularities, with their relative contributions depending on the representation, architecture, training objective, and data regime[6,8,11,20,21]. RAVEN adopts a narrower architectural intervention by fixing the graph encoders before affinity fitting, so affinity labels cannot directly update the structural projection itself. This design does not remove bias from the supervised problem, but it separates task-independent structural feature generation from downstream affinity regression and eliminates one route by which the affinity objective could reshape encoder parameters. This separation motivates the study of frozen graph projections whose predictive utility is recovered through comparatively lightweight supervised readers.

The use of fixed random graph encoders has precedents in reservoir computing and random-feature learning[25–28]. Graph Echo State Networks separate an untrained random graph reservoir from an adaptive readout, while related reservoir analyses connect randomly fixed internal representations with kernel limits[25,28]. Graph Random Neural Features extend the random-feature viewpoint to graph-structured inputs by constructing explicit embeddings from independently sampled neural mappings[26]. More generally, random-feature methods interpret independently sampled feature maps as finite approximations to similarities induced by an underlying feature distribution[26,27]. This perspective provides a basis for treating multiple independently initialized frozen graph encoders as finite samples from an initialization-induced distribution over structural feature maps. RAVEN develops this principle through a multi-seed atomistic graph reservoir and a structurally asymmetric global encoder, allowing distinct frozen structural projections to coexist within one representation space. A deterministic physicochemical fingerprint forms a parallel representation pathway, supplying chemically defined interaction statistics that need not be represented consistently by any individual random projection[15,19]. Neural and tree-based readers then impose different supervised inductive biases on the frozen structural and deterministic physicochemical representations, while validation-based fusion combines predictions arising from these heterogeneous model families. Related affinity ensemble and meta-modeling studies have shown that predictors based on different representations or model families can provide complementary information when their outputs are integrated[29,30]. RAVEN consequently treats diversity as an architectural resource spanning independently sampled graph encoders, structural and physicochemical representations, and heterogeneous downstream readers.

## Methods

### Framework Overview

RAVEN integrates an atomistic heterogeneous graph representation, a frozen multi-view graph reservoir, a deterministic physicochemical fingerprint, and heterogeneous supervised regressors. The complete workflow is summarized in Figure 1. Each protein–ligand complex is represented as a heterogeneous graph containing ligand and protein atoms, intramolecular covalent relations, and directed cross-interface contacts. The graph provides a shared structural input to two complementary representation pathways. The first pathway consists of multiple independently initialized relation-specific graph encoders. Each encoder generates a protein–interaction view and a ligand view, while an additional independently initialized encoder supplies an asymmetric global protein–ligand branch. All graph-encoder parameters remain fixed after initialization, preventing affinity supervision from modifying the structural projections. The second pathway constructs a deterministic 788-dimensional physicochemical fingerprint from graph composition, atom and bond statistics, cross-interface geometry, and channel-resolved in-teraction annotations. This representation preserves explicit structural and physicochemical information that may not be expressed consistently across individual random graph projections. Three supervised expert families operate on these representations. Reservoir MLPs use the frozen structural representation alone, joint MLPs combine the structural reservoir with the physicochemical fingerprint, and ExtraTrees regressors operate directly on the deterministic fingerprint. Three independently trained experts are used within each family. Their predictions are combined through a nonnegative softmax-weighted fusion rule fitted on the validation partition and fixed before evaluation on the held-out test partition. The architecture therefore introduces complementary variation at three levels: independent random graph realizations, structural and physicochemical representation channels, and neural and tree-based supervised readers. Subsequent sections describe the construction and operation of each component.

**Atomistic Heterogeneous Graph Representation**

Each protein–ligand complex was represented as a directed atomistic heterogeneous graph, as illustrated in Figure 1A. Ligand atoms and protein-pocket atoms formed two distinct node types, while intramolecular covalent bonds and intermolecular contacts were represented as separate relation types. This organization preserves the molecular identities of the ligand and protein domains while exposing their internal connectivity and cross-interface interactions independently to the graph encoder.

Ligand structures were parsed using RDKit. Explicit hydrogen atoms were omitted from the graph, while the total number of attached hydrogens was retained as an atomic attribute. Protein pockets were represented at full heavy-atom resolution. Invariant atomic attributes and intraresidue covalent topologies were cached in residue-level heavy-atom templates to accelerate preprocessing and maintain consistent bond perception. Observed Cartesian coordinates were assigned by atom-name matching, and bidirectional peptide bonds were added between consecutive residues. The template library therefore served only as a deterministic preprocessing mechanism and did not alter the atom-level resolution of the resulting protein graph.

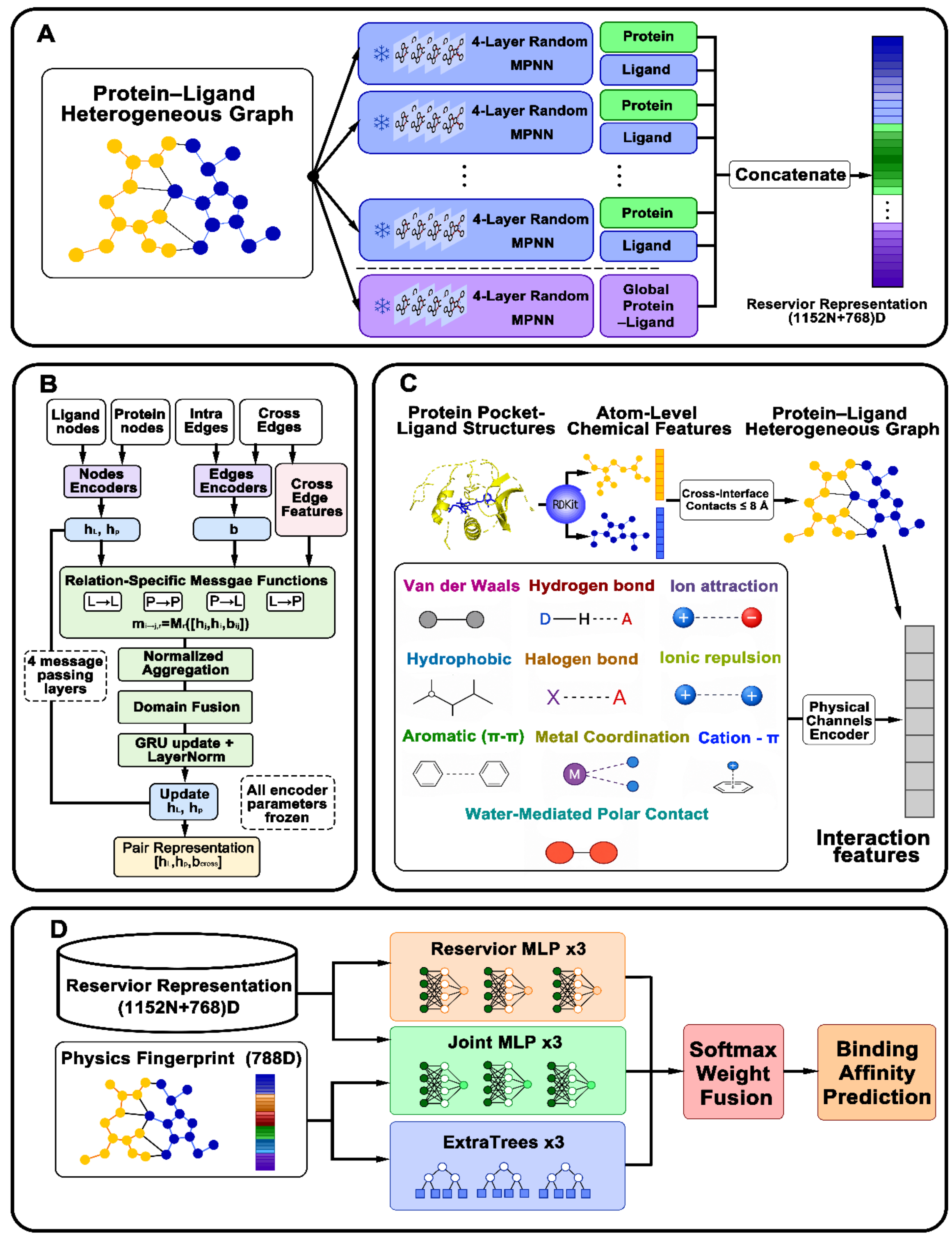

**Figure 1.** Overview of RAVEN. (A) Multiple independently initialized frozen graph encoders and an additional asymmetric global readout branch generate complementary structural views. (B) Each encoder performs relation-specific heterogeneous message passing with square-root degree-normalized aggregation and gated node-state updates. (C) Protein and ligand structures are converted into an atomistic heterogeneous graph containing intramolecular and cross-interface relations. (D) The frozen structural reservoir and deterministic 788-dimensional physico-

chemical fingerprint are processed by structure-only MLPs, joint structure–physics MLPs, and physics-only ExtraTrees regressors, followed by validation-fitted nonnegative fusion.

Ligand and protein atoms shared a 41-dimensional chemical feature space comprising element identity, atomic degree, formal charge, radical-electron count, hybridization state, aromaticity, total hydrogen count, and chirality. The element encoding distinguished C, N, O, S, F, P, Cl, Br, I, B, Si, Fe, Zn, Cu, Mn, Mo, and other elements. Intramolecular covalent edges were described by 10-dimensional attributes encoding bond order, conjugation, ring membership, and bond stereochemistry, and every covalent connection was stored in both directions. Protein intramolecular edges included both residue-template covalent bonds and peptide bonds between consecutive residues. The resulting heterograph contained four directed relation types, denoted $E_{LL}$, $E_{PP}$, $E_{LP}$, and $E_{PL}$, corresponding to ligand intramolecular bonds, protein intramolecular bonds, and the two directions of protein–ligand cross-interface contacts, respectively. The two intermolecular relations shared the same atom-pair annotations but used reversed source and destination assignments, allowing direction-specific message functions during graph propagation. Related atomistic graph models have similarly represented molecular complexes using atomic properties, covalent topology, and spatially defined intermolecular neighborhoods[16].

Cross-interface connectivity was constructed directly from the three-dimensional atomic coordinates. For each ligand atom, up to 32 nearest protein atoms were considered, and pairs separated by no more than 8 Å were retained. When more than 4096 cross-interface pairs were available for a complex, the shortest-distance pairs were preserved. Pairs within 6 Å were additionally identified as direct contacts, whereas pairs between 6 and 8 Å provided broader local structural context. For every retained ligand–protein pair, the model recorded the interatomic distance and directed displacement together with ten binary physicochemical interaction gates and ten continuous geometry-aware quality scores. These annotations indicate whether an atom pair satisfies a channel-specific interaction criterion and quantify its agreement with the corresponding distance and orientation preferences; their detailed chemical definitions are given in the physicochemical-fingerprint subsection. The four relation types were processed by relation-specific message functions, after which ligand-node, protein-node, and cross-interface pair states were summarized by mean and maximum pooling following four propagation layers. The detailed propagation and frozen multi-encoder formulation are described in the following subsection within the general message-passing framework[31]. Graph storage, batching, and relation-aware tensor operations were implemented using DGL[32].

### Frozen Multi-View Graph Reservoir and Random-Feature Analysis

The atomistic heterograph was processed by multiple independently initialized relation-specific message-passing neural networks. Each encoder contained four propagation layers and remained fully frozen after initialization, preserving the separation between graph representation and supervised affinity regression used in graph-reservoir formulations[25,31]. A random encoder realization, denoted by $\omega \sim P_{\Omega}$, refers to one complete parameterized MPNN including the ligand and protein node encoders, intramolecular and cross-interface edge encoders, all relation-specific message functions, recurrent node-update modules, normalization parameters, and the final cross-interface pair network. The Monte Carlo unit considered below is therefore an entire independently initialized graph encoder rather than an individual hidden coordinate.

Ligand and protein node features were first mapped to 192-dimensional latent states, whereas intramolecular and cross-interface edge descriptors were mapped to 96-dimensional embeddings. At propagation layer $\ell$, each of the four relation types used an independently parameterized message function. Relation-specific messages were aggregated by a square-root degree-normalized sum and subsequently merged according to the molecular domain of the target node. The resulting aggregate updated the corresponding ligand or protein state through a domain-specific GRUCell followed by LayerNorm. The principal propagation step can be summarized as

$$\mathbf{m}_{j\to i,r}^{(\ell)} = M_r^{(\ell)}\left(\left[\mathbf{h}_j^{(\ell)};\mathbf{h}_i^{(\ell)};\mathbf{b}_{ji}^{r}\right]\right), \quad (1)$$

$$\mathbf{a}_{i,r}^{(\ell)} = \frac{\sum_{j\in\mathcal{N}_r(i)}\mathbf{m}_{j\to i,r}^{(\ell)}}{\sqrt{\max(|\mathcal{N}_r(i)|,1)}}, \quad (2)$$

$$\mathbf{h}_i^{D,(\ell+1)} = \mathrm{LN}_D^{(\ell)}\left[\mathrm{GRU}_D^{(\ell)}\left(\sum_{r\in\mathcal{R}_D}\mathbf{a}_{i,r}^{(\ell)}, \mathbf{h}_i^{D,(\ell)}\right)\right], \quad (3)$$

where $D \in \{L, P\}$, $\mathcal{R}_L = \{LL, PL\}$, and $\mathcal{R}_P = \{PP, LP\}$. Parameters were not shared across propagation depths. After four layers, an additional pair network generated cross-interface interaction states from the final ligand and protein states together with the corresponding edge embedding. Mean and maximum pooling were then applied independently to protein nodes, ligand nodes, and cross-interface pair states. The complete layerwise transformations, dimensional specifications, edge encoders, pair-network definition, and pooling operations are provided in Appendix.

Each complete ordinary encoder realization consequently produced three 384-dimensional graph-level views representing the protein, cross-interface interaction, and ligand domains. Its random graph feature block was therefore defined as

$$\boldsymbol{\phi}_\omega(G) = [\mathbf{p}_\omega(G); \mathbf{i}_\omega(G); \mathbf{l}_\omega(G)] \in \mathbb{R}^{1152}. \tag{4}$$

For $M$ independently initialized ordinary encoders, $\omega_1, \dots, \omega_M \overset{\text{i.i.d.}}{\sim} P_\Omega$, these blocks were preserved by concatenation rather than averaging. An additional independently initialized encoder used the same underlying MPNN architecture but supplied an asymmetric global protein–ligand readout $\boldsymbol{\gamma}_\nu(G) \in \mathbb{R}^{768}$. The complete unadapted structural reservoir was

$$\mathbf{R}_M(G) = \left[\boldsymbol{\phi}_{\omega_1}(G); \cdots; \boldsymbol{\phi}_{\omega_M}(G); \boldsymbol{\gamma}_\nu(G)\right] \in \mathbb{R}^{1152M+768}. \tag{5}$$

The final configuration used $M = 32$, yielding a 37,632-dimensional raw structural representation. Because the global branch uses a different graph-level readout, it is treated as a complementary auxiliary branch rather than as an additional identically distributed ordinary encoder realization.

The ordinary encoder bank admits a network-level random-feature interpretation related to random kernel approximation and graph random neural features[26,27]. Let $\mathbf{C}_M(G) = \left[\boldsymbol{\phi}_{\omega_1}(G); \cdots; \boldsymbol{\phi}_{\omega_M}(G)\right]$ denote the concatenated ordinary-head reservoir. For two heterographs $G$ and $G'$, the scale-normalized inner product induced by this concatenated representation is

$$\widehat{K}_M(G, G') = \frac{1}{M}\mathbf{C}_M(G)^\top \mathbf{C}_M(G') = \frac{1}{M}\sum_{m=1}^{M} \boldsymbol{\phi}_{\omega_m}(G)^\top \boldsymbol{\phi}_{\omega_m}(G'). \tag{6}$$

Thus, the implemented model retains the individual encoder blocks by concatenation, while the scale-normalized similarity induced by that representation is exactly the arithmetic mean of the similarities contributed by the independent encoder realizations. For a single realization, define $X_\omega(G, G') = \boldsymbol{\phi}_\omega(G)^\top \boldsymbol{\phi}_\omega(G')$ and $K_\Omega(G, G') = \mathbb{E}_\omega[X_\omega(G, G')]$.

**Suggest a viewpoint.** Assume that the complete ordinary-encoder realizations are independent and identically distributed and that $X_\omega(G, G')$ has finite variance $\sigma^2_{G,G'}$. Then

$$\mathbb{E}\left[\widehat{K}_M(G, G')\right] = K_\Omega(G, G'), \qquad \mathrm{Var}\left[\widehat{K}_M(G, G')\right] = \frac{\sigma^2_{G,G'}}{M}, \qquad \widehat{K}_M(G, G') - K_\Omega(G, G') = O_p\left(M^{-1/2}\right). \tag{7}$$

The viewpoint establishes that increasing the number of independent complete encoder realizations reduces the initialization variance of the scale-normalized structural similarity induced by the ordinary reservoir. The justification, finite-sample concentration bounds, the corresponding feature-distance result, the representation-diversity identity, and the distinction between independent encoder replication and conventional within-network widening are given in Appendix.

All graph-encoder parameters remained fixed throughout affinity regression, so supervised gradients were restricted to the downstream readers and could not reshape the sampled atomistic projections. The concentration result concerns initialization dependence of the induced structural similarity and does not imply that the downstream model averages the concatenated feature blocks or that predictive error must decrease monotonically with $M$. In finite samples, increasing $M$ simultaneously enlarges the supervised representation dimension, and predictive risk may therefore remain nonmonotonic as sample size, regularization, and reader capacity interact with reservoir size[33]. The final reservoir size was consequently selected empirically using the Validation partition.

**Deterministic Physicochemical Interaction Fingerprint**

The deterministic physicochemical branch was designed as a parallel representation pathway complementary to the frozen random graph reservoir. Whereas the random encoders describe atomistic structure through initialization-dependent nonlinear projections, the deterministic branch preserves chemically named composition, geometry, and interaction statistics whose definitions remain fixed across encoder realizations. This separation allows the supervised readers to combine structurally expressive random features with reproducible physicochemical information rather than requiring either representation family to encode the complete binding interface alone. Fixed interaction fingerprints, chemically resolved atom-pair representations, multiscale contact descriptors, and hybrid physics–learning architectures provide precedents for this representation strategy[15,18,19,34–37].

For each protein–ligand heterograph $G$, the deterministic mapping produced a fixed-length physicochemical fingerprint

$$\mathbf{f}_{\text{phys}}(G) \in \mathbb{R}^{788}. \tag{8}$$

Its construction used no affinity labels, trainable parameters, or random sampling, such that the same encoded complex produced the same physicochemical representation. The fingerprint was organized into three complementary information levels: molecular composition and intramolecular topology, cross-interface geometry, and channel-resolved physicochemical interaction statistics.

Atomic chemical roles were assigned deterministically from the graph features and local molecular connectivity. These assignments distinguished hydrogen-bond donors and acceptors, positively and negatively charged atoms, hydrophobic atoms, halogen donors, metals, and polar atoms. They were subsequently used to identify ten physicochemical interaction categories: van der Waals contact, hydrogen bonding, ionic attraction, ionic repulsion, hydrophobic contact, aromatic interaction, metal coordination, halogen bonding, cation–$\pi$ interaction, and a putative wa-ter-mediated polar-contact proxy. For each category, the representation retained both interaction occurrence and geometry-aware quality information. These channels are physics-informed interaction categories rather than an additive decomposition of binding free energy. Their purpose is to expose chemically interpretable interaction patterns while leaving the statistical relationship between those patterns and experimental affinity to the supervised readers.

Variable-size molecular and interface information was converted into a fixed-length representation through multiscale statistical summaries. The first 522 dimensions describe molecular composition and intramolecular topology, including graph-level size and density descriptors together with separate ligand and protein node and bond statistics. Keeping the two molecular domains separate preserves their asymmetric chemical roles and avoids allowing the substantially larger protein pocket to numerically obscure ligand-specific information.

A further 64 dimensions describe the overall cross-interface geometry. These features summarize interatomic distance distributions, coarse pair-direction information, and overlapping radial-basis representations of intermolecular separation. The combination of discrete distance occupancy and smooth radial representations retains both local contact density and gradual geometric variation across neighboring distance ranges.

The remaining 202 dimensions summarize the occurrence, spatial distribution, geometric quality, and diversity of the ten physicochemical interaction channels. The representation includes absolute and interface-normalized interaction prevalence, distance-sensitive event summaries, active-contact distance statistics, interaction-quality summaries, and global interaction-diversity descriptors. The complete composition of the 788-dimensional fingerprint is summarized in Table 1.

**Table 1.** Composition of the deterministic 788-dimensional physicochemical interaction fingerprint.

| Feature group | Dim. | Information retained |
|---|---|---|
| Global graph size and density | 12 | Molecular size, intramolecular connectivity, cross-interface occupancy, and normalized contact density. |
| Ligand node statistics | 205 | Five statistical summaries of each of the 41 ligand atom features. |
| Protein node statistics | 205 | Five statistical summaries of each of the 41 protein atom features. |
| Ligand bond statistics | 50 | Five statistical summaries of the ten ligand intramolecular edge features. |
| Protein bond statistics | 50 | Five statistical summaries of the ten protein intramolecular edge features. |
| Cross-distance statistics | 5 | Mean, standard deviation, minimum, maximum, and square-root-normalized sum of cross-interface distances. |
| Distance histogram | 24 | Logarithmic counts and normalized ratios across 12 intermolecular distance intervals. |
| Cartesian direction statistics | 9 | Mean, standard deviation, and mean absolute value of the three pair-direction coordinates. |
| Radial-basis statistics | 26 | Mean and square-root-normalized sum for 13 overlapping radial distance bases. |
| Channel occurrence statistics | 30 | Counts, interface ratios, and square-root-normalized counts for the ten physicochemical channels. |
| Distance-weighted channel events | 40 | Four distance-decay summaries for every interaction channel. |
| Active-channel distance statistics | 50 | Event count and four distance-distribution summaries for every interaction channel. |

| Feature group | Dim. | Information retained |
|---|---|---|
| Interaction-quality statistics | 50 | Five statistical summaries of each continuous channel-quality coordinate. |
| Active-quality statistics | 30 | Mean, cumulative, and maximum quality among active interaction events. |
| Channel diversity and event total | 2 | Number of represented interaction types and total gated interaction activity. |
| Total | 788 | Deterministic multiscale description of molecular composition, interface geometry, and physicochemical interaction patterns. |

The fingerprint intentionally retains partially overlapping descriptors across multiple spatial and statistical scales rather than imposing a single predefined summary scale or applying post hoc feature elimination. This design allows neural and tree-based readers to select different projections of the same molecular interface according to their respective inductive biases. All distance-based, ring-geometric, and interaction-count features are invariant to rigid translation, and all except the nine Cartesian pair-direction summaries are also invariant to rigid rotation. The resulting 788-dimensional fingerprint should therefore be interpreted as an explicit and reproducible physicochemical description of the encoded complex rather than as an exact mechanistic or thermodynamic model of binding affinity.

**Heterogeneous Regressors and Nonnegative Fusion**

The frozen structural reservoir and deterministic physicochemical fingerprint were decoded by three complementary supervised model families: reservoir-only multilayer perceptrons (R-MLPs), physics-only Extremely Randomized Trees (ExtraTrees), and joint structure–physics multilayer perceptrons (J-MLPs). Three independently initialized instances were trained within each family, yielding nine candidate experts for prediction-level fusion. This heterogeneous readout design extends representation diversity into the supervised stage by combining different information sources and function classes with partially distinct inductive biases[29].

The raw structural reservoir grows rapidly with the number of frozen encoder realizations and contains distinct protein–interaction, ligand, and global representation branches. Rather than directly supplying this high-dimensional concatenation to the downstream neural regressor, the branches were compressed independently using lightweight trainable adapters and concatenated only after dimensionality reduction. No cross-branch interaction was introduced within the adapter stage. The resulting compact representation was used by the R-MLP as a structure-only supervised pathway, without access to the deterministic physicochemical fingerprint. This reader therefore measures the predictive information that can be extracted from the frozen random structural reservoir alone.

The second supervised family consisted of physics-only ExtraTrees regressors operating directly on the unstandardized 788-dimensional physicochemical fingerprint. The fingerprint contains heterogeneous descriptors including interaction counts, distance distributions, extrema, geometric-quality summaries, and multiscale interface statistics. Tree-based threshold partitioning provides a substantially different inductive bias from the smooth distributed mappings learned by neural readers and can naturally represent conditional feature combinations and locally defined response regions[38]. Because individual tree splits depend on the ordering of values within a feature rather than on distances across differently scaled coordinates, the original physicochemical features were retained without numerical standardization. This pathway provides a supervised estimate based exclusively on explicit physicochemical information and remains independent of the frozen graph reservoir.

The J-MLP connected the two representation pathways at the feature level. It combined the adapted structural reservoir with a compact neural projection of the deterministic physicochemical fingerprint before nonlinear regression. This joint representation allows the supervised reader to model dependencies between randomized atomistic structural projections and explicit interaction statistics before forming the final affinity estimate. Accordingly, the comparison between R-MLP and J-MLP readers provides a direct assessment of the additional predictive information contributed by the physicochemical representation beyond the frozen structural reservoir, while the physics-only ExtraTrees pathway provides a complementary function class operating on the same deterministic information source.

The resulting expert pool therefore comprised three conceptually distinct prediction pathways: structure-only neural regression, physics-only tree regression, and joint structure–physics neural regression. Their complementarity arises from both the information supplied to each reader and the statistical form of the corresponding supervised mapping. Detailed adapter architectures, neural layer dimensions, regularization settings, tree hyperparameters, and optimization procedures are provided in the Supporting Information.

The nine expert predictions were combined through a validation-fitted nonnegative softmax weighting scheme,

$$\hat{y}_{\mathrm{ens}}(G) = \sum_{k=1}^{9} w_k\, \hat{y}_k(G), \qquad w_k \geq 0, \qquad \sum_{k=1}^{9} w_k = 1. \tag{9}$$

All base-expert parameters remained fixed during this stage. The fusion weights were optimized only on the Validation partition with weak regularization and were subsequently fixed before evaluation on the Test partition and the protected CASF-2016 subset. This strategy follows the general motivation of stacked generalization while deliberately restricting the second-level predictor to an interpretable convex combination rather than an unrestricted meta-regressor[39].

**Data Partitioning, Training, and Evaluation**

The study used the PDBbind 2020R1 protein–ligand complexes that remained available after structure cleaning and successful heterogeneous-graph construction[40]. All experimental affinity labels, model predictions, and error metrics were expressed on the dimensionless pK scale.

A mutually isolated split was reconstructed for the processed RAVEN dataset using the pairwise similarity matrices released with the GEMS project[41,42]. The published GEMS CleanSplit member list was not used directly because the cleaning and graph-construction procedures produced a project-specific set of successfully processed complexes. Instead, the GEMS similarity resources and conflict criteria were applied to the RAVEN sample manifest to reconstruct Train, Validation, and Test membership. The final processed dataset contained 11,657 protein–ligand complexes, comprising 8,161 Train complexes, 1,747 Validation complexes, and 1,749 Test complexes. The Test partition included all 285 available CASF-2016 complexes as a protected subset. GEMS similarity records were matched to the processed complexes using normalized PDB identifiers.

For complexes $i$ and $j$, a similarity-conflict indicator was defined as

$$C_{ij} = \mathbb{I}\left(\left|\mathrm{pK}_i - \mathrm{pK}_j\right| < 1.0\right)\mathbb{I}\left[T_{ij} > 0.9 \;\vee\; \left(\mathrm{TM}_{ij} > 0.8 \;\wedge\; T_{ij} + \left(1 - \mathrm{RMSD}_{ij}\right) > 0.8\right)\right], \tag{10}$$

where $T_{ij}$, $\mathrm{TM}_{ij}$, and $\mathrm{RMSD}_{ij}$ denote ligand Tanimoto similarity, protein TM-score, and pocket-aligned ligand RMSD, respectively. A pair was therefore considered conflicting when its affinity difference was smaller than 1.0 pK unit and it exhibited either strong ligand similarity or combined protein-structure and binding-conformation similarity. These quantities characterize overlap at the ligand, protein, and bound-complex levels and follow the data-bias analysis underlying the GEMS resources[41].

All conflicting pairs were represented as edges of an undirected graph whose vertices corresponded to successfully processed complexes. Each connected component was treated as an indivisible allocation unit, preventing complexes connected through either direct or transitive similarity conflicts from being separated across data subsets. All available CASF-2016 complexes, together with every additional complex belonging to the same similarity-connected components, were assigned to Test[43]. The remaining components were allocated among Train, Validation, and Test with approximate target proportions of 70%, 15%, and 15%, using affinity-stratified assignment to preserve the pK distribution as closely as possible. Because connected components varied in size and could not be divided, the realized proportions were allowed to differ slightly from these nominal targets.

For any two distinct subsets $A$ and $B$, the resulting isolation condition was

$$\sum_{i\in A}\sum_{j\in B} C_{ij} = 0, \qquad A, B \in \{\text{Train}, \text{Validation}, \text{Test}\}, \qquad A \neq B. \tag{11}$$

This condition states that no pair drawn from different subsets satisfies the predefined similarity-conflict criterion. It does not imply complete unrelatedness under every possible sequence-, topology-, ligand-, or structure-similarity definition.

Reservoir-size experiments evaluated $M \in \{1,2,3,4,6,8,10,12,14,16,20,32\}$ ordinary random encoders. A single ordered bank of independently initialized encoders was generated once, and each reservoir size used a nested prefix of this bank together with the same separately initialized global protein–ligand branch. This construction ensured that smaller reservoirs were contained within larger reservoirs rather than being regenerated from unrelated initialization sets. The final reservoir size was selected using Validation performance only.

After data partitioning, all preprocessing statistics and supervised parameters were fitted exclusively on Train. For the neural readers, structural-reservoir and physicochemical-fingerprint coordinates were standardized using Train-derived means and standard deviations. The pK targets were standardized in the same manner during neural optimization, and predictions were transformed back to the original pK scale before evaluation. The Train-derived transformations were fixed when applied to Validation and Test. ExtraTrees instead received the original unstandardized physicochemical fingerprint and predicted directly on the pK scale.

All random graph encoders remained frozen and in inference mode throughout supervised training. Affinity gradients were restricted to the trainable adapters, branch-weight parameters, and regression heads of the neural readers. The neural experts were optimized with AdamW and included weight decay, Gaussian perturbation of standardized inputs, gradient clipping, learning-rate reduction on Validation plateaus, and early stopping. Detailed optimizer settings and regularization hyperparameters are provided in the Supporting Information.

Neural optimization combined robust regression with a minibatch correlation objective,

$$\mathcal{L}_{\text{neural}} = \frac{1}{B}\sum_{i=1}^{B} \ell_{0.5}\left(\hat{\tilde{y}}_i - \tilde{y}_i\right) + 0.08\left[1 - r_B\left(\hat{\tilde{\mathbf{y}}}, \tilde{\mathbf{y}}\right)\right], \tag{12}$$

where $\ell_{0.5}$ is the SmoothL1 loss with transition parameter $0.5$, $B$ is the minibatch size, and $r_B$ denotes Pearson correlation within the minibatch. The first term provides a robust penalty on prediction error, whereas the second encourages preservation of affinity variation across samples. For each independently initialized neural expert, the state with the highest Validation Pearson correlation was retained. ExtraTrees experts were fitted once on Train and required no epoch-wise checkpoint selection.

After all nine base experts had been trained and fixed, the nonnegative fusion weights were optimized using Validation predictions only. With $\mathbf{a} \in \mathbb{R}^9$ denoting the fusion logits and $\mathbf{w} = \mathrm{softmax}(\mathbf{a})$, the fusion objective was

$$\mathcal{L}_{\text{fusion}} = 1 - r\left(\sum_{k=1}^{9} w_k \, \hat{\mathbf{y}}_k^{\text{Validation}}, \mathbf{y}^{\text{Validation}}\right) + 0.015\|\mathbf{a}\|_2^2. \tag{13}$$

All base-expert parameters remained unchanged during this stage. Validation was therefore used for neural checkpoint selection, reservoir-size selection, and fusion-weight fitting, whereas Test was excluded from all parameter fitting and model-selection decisions[44]. The resulting preprocessing transformations, expert states, and fusion weights were fixed before final evaluation. CASF-2016 performance was reported separately from the complete Test results, although the available CASF complexes remained part of the protected Test allocation.

Predictive performance was evaluated on the original dimensionless pK scale using Pearson correlation $r$, Spearman rank correlation $\rho$, root-mean-square error (RMSE), mean absolute error (MAE), mean prediction bias, and the coefficient of determination $R^2$. Readout latency was measured after structural and physicochemical representation construction using three warm-up repetitions followed by ten timed repetitions and was normalized by the number of evaluated complexes. The neural models were implemented in PyTorch, whereas the ExtraTrees regressors and conventional evaluation utilities were implemented using scikit-learn[45,46].

# Results

## Overall Predictive Performance of the Final 32-Head Model

The final reported RAVEN model used 32 ordinary frozen graph encoders together with one additional global protein–ligand branch. The resulting structural reservoir contained 33 parallel representation branches and produced a 37,632-dimensional raw feature vector. Three R-MLPs, three J-MLPs, and three ExtraTrees regressors formed the nine-expert prediction pool, whose outputs were combined through nonnegative softmax fusion. Table 2 summarizes the predictive performance of this model on the Validation partition, the complete PDBbind 2020R1 Test set, and the protected CASF-2016 subset. Because the fusion weights were fitted on Validation predictions, the Validation metrics are included as descriptive results, whereas the Test and CASF-2016 results characterize held-out predictive performance.

**Table 2.** Predictive performance of the final 32-head RAVEN model. RMSE, MAE, and Bias are reported on the dimensionless pK scale. Validation metrics are descriptive because the nonnegative fusion weights were fitted using this partition. CASF-2016 is a protected subset of the complete PDBbind 2020R1 Test set.

| Dataset | $N$ | Pearson $r$ | Spearman $\rho$ | $R^2$ | RMSE | MAE | Bias |
|---|---|---|---|---|---|---|---|
| Validation | 1747 | 0.7795 | 0.7722 | 0.6076 | 1.1851 | 0.8892 | −0.0004 |
| PDBbind 2020R1 Test | 1749 | 0.7995 | 0.7922 | 0.6341 | 1.2396 | 0.9379 | −0.0483 |
| CASF-2016 subset | 285 | 0.8446 | 0.8407 | 0.6895 | 1.2095 | 0.9243 | −0.0072 |

On the complete PDBbind 2020R1 Test set, the 32-head model achieved a Pearson correlation of 0.7995, a Spearman correlation of 0.7922, and an $R^2$ of 0.6341. The corresponding RMSE and MAE were 1.2396 and 0.9379 on the dimensionless pK scale, respectively. The prediction errors had a mean of −0.048, a standard deviation of 1.234, and a median of −0.040.

Within the protected CASF-2016 subset43, Pearson (r) and Spearman ($\rho$) reached 0.8446 and 0.8407, respectively, while $R^2$ reached 0.6895. The corresponding RMSE and MAE were 1.2095 and 0.9243. The prediction errors had a mean of 0.007, a standard deviation of 1.220, and a median of −0.045. The MAE remained below 1.0 pK in both held-out evaluations, while the correlation coefficients showed that the model retained continuous affinity variation and rank ordering across the evaluated complexes.

As shown in Figure 2, the error distributions for both held-out evaluation sets were centered close to zero. On the complete PDBbind 2020R1 Test set, the mean and median errors were −0.048 and −0.040, respectively, with a standard deviation of 1.234. On the CASF-2016 subset, the corresponding mean and median errors were 0.007 and −0.045, with a standard deviation of 1.220. The similar dispersion observed across the two evaluation sets indicates that the prediction-error scale remained broadly consistent between the complete Test set and the protected CASF-2016 subset.

**Reservoir-Size Scaling and the Contribution of the Global Protein–Ligand Branch**

The effect of reservoir size was evaluated by varying the number of ordinary frozen random encoders while retaining the same supervised-reader definitions and training protocol. The evaluated head counts were $M \in \{1,2,3,4,6,8,10,12,14,16,20,32\}$. Each reservoir was constructed as a nested prefix of the same ordered encoder bank, ensuring that every smaller reservoir was con-tained within the larger configurations. At each value of $M$, the complete nine-expert model and the structure-only R-MLP path were evaluated both with and without the additional global protein–ligand branch. The R-MLP-only results represent the arithmetic mean of three independently initialized R-MLP experts. Figure 3 summarizes the resulting Pearson correlations and RMSE values on the complete PDBbind 2020R1 Test set and the protected CASF-2016 subset.

The reservoir-size effect was most clearly expressed in the structure-only R-MLP path. With the global branch retained, increasing $M$ from 1 to 32 raised the mean PDBbind Test Pearson correlation from 0.7548 to 0.7803 and reduced RMSE from 1.3486 to 1.2864 pK. On CASF-2016, the corresponding Pearson correlation increased from 0.7976 to 0.8172, while RMSE decreased from 1.3496 to 1.2523 pK. The same overall scaling direction remained after removal of the global branch. Under this ablation, the PDBbind Test Pearson correlation increased from 0.7320 to 0.7712 and RMSE decreased from 1.3872 to 1.2956 pK, whereas the CASF-2016 Pearson correlation increased from 0.7550 to 0.8091 and RMSE decreased from 1.3723 to 1.2806 pK. The agreement between the two evaluation sets and the two global-branch conditions indicates that increasing the number of independent frozen projections improved the supervised readability of the structure-only representation on average.

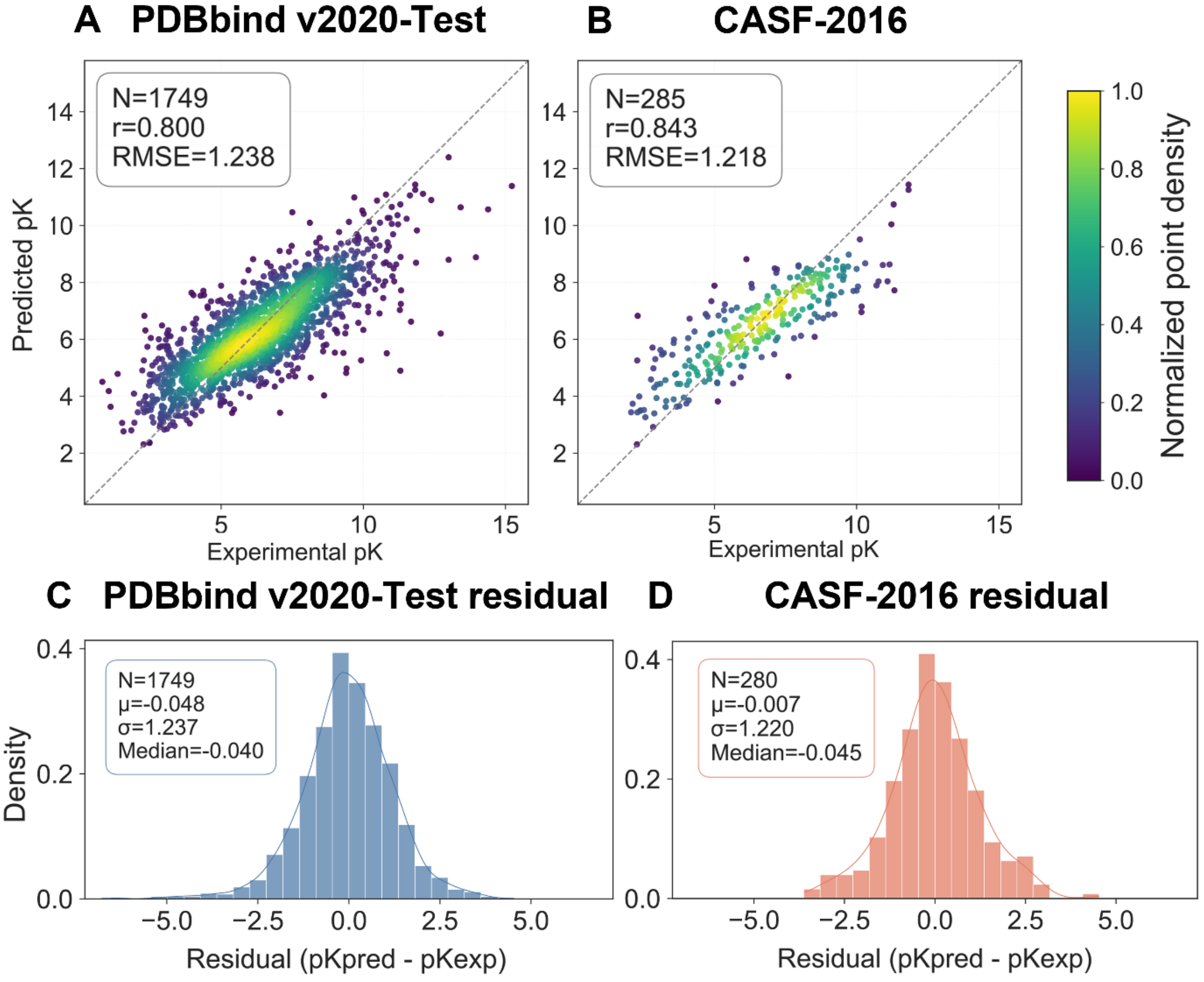

**Figure 2.** Prediction and residual analysis for the final 32-head RAVEN model. Panels (A) and (B) show predicted versus experimental affinities for the complete PDBbind 2020R1 Test set and the protected CASF-2016 subset, respectively. Point colors represent normalized local sample density, and the dashed diagonal indicates ideal agreement between predicted and experimental values. Panels (C) and (D) show the corresponding residual distributions, where each residual is defined as predicted pK minus experimental pK.All affinities and error metrics are reported on the dimensionless pK scale.

The largest R-MLP gains occurred between the single-head configuration and the low-to-moderate reservoir sizes, followed by a comparatively stable region at approximately $M = 12$–16 and above. The curves nevertheless contained local reversals, particularly among the smaller reservoirs, and therefore did not support a strictly monotonic relationship between head count and predictive performance. At small $M$, each additional encoder constitutes a relatively large change to the finite random-feature pool, and its projection may contribute a new structural direction or partially overlap with information already represented by the existing heads.

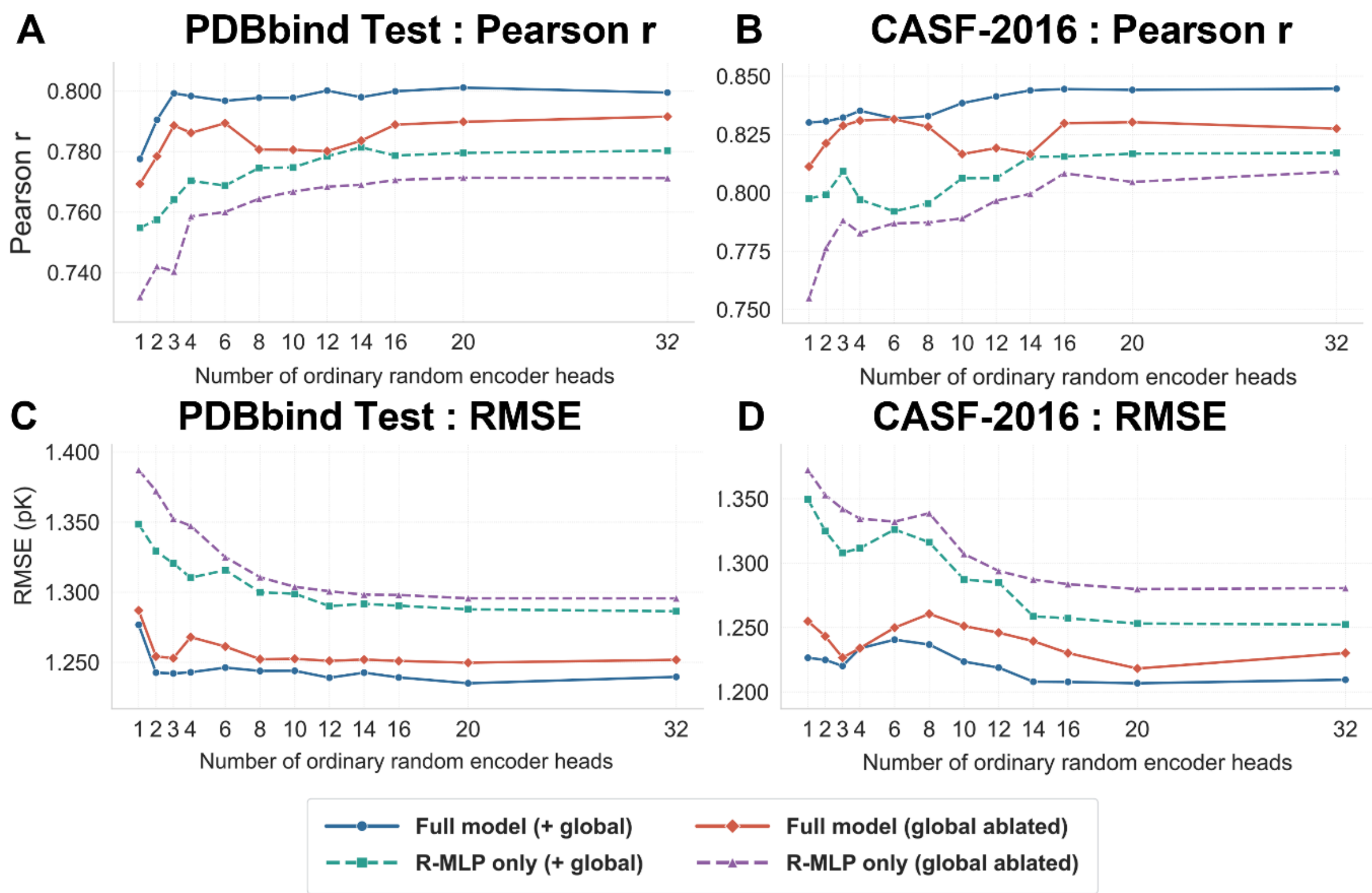

**Figure 3.** Effects of reservoir size and the global protein–ligand branch. Panels (A) and (B) show Pearson correlation on the complete PDBbind 2020R1 Test set and the protected CASF-2016 subset, respectively, while panels (C) and (D) show the corresponding RMSE values on the dimensionless pK scale. The ordinary-head count $M$ excludes the separately initialized global branch. Full-model curves represent the nonnegative softmax fusion of three R-MLPs, three J-MLPs, and three ExtraTrees experts. R-MLP-only values represent the arithmetic mean of three independently initialized R-MLP experts at each reservoir size. Solid curves correspond to the complete expert ensemble, while dashed curves correspond to the structure-only R-MLP family. Circle and square markers indicate models retaining the global branch, whereas diamond and triangle markers indicate the corresponding global-branch ablations. Higher Pearson correlation and lower RMSE indicate better predictive performance.

Increasing $M$ also enlarges the adapted R-MLP input from $384M + 256$ dimensions, corresponding to 640 dimensions at $M = 1$ and 12,544 dimensions at $M = 32$, while the number of training complexes remains unchanged. The gain in structural coverage may therefore be partially opposed by the higher-dimensional supervised estimation problem, including increased estimation variance and greater sensitivity to finite-sample fitting. This competition between representation expansion and supervised estimation provides a plausible basis for the observed local nonmonotonicity and diminishing returns, consistent with nonmonotonic risk behavior reported for multiple-random-feature models.[33]

The dependence on reservoir size was less pronounced in the complete expert ensemble. On the PDBbind Test set, Pearson correlation increased from 0.7776 at $M = 1$ to 0.7993 at $M = 3$, after which it remained within the narrow range of 0.7968–0.8011 across the remaining reservoir sizes. The corresponding RMSE decreased from 1.2768 to 1.2421 pK by $M = 3$ and subsequently remained between 1.2351 and 1.2462 pK. On CASF-2016, the complete model improved more gradually, with Pearson correlation increasing from 0.8302 at $M = 1$ to 0.8446 at $M = 32$, while RMSE decreased from 1.2265 to 1.2095 pK. The smaller head-count dependence of the complete model indicates that its deterministic physicochemical pathway and heterogeneous supervised readers supplied a strong predictive baseline before the random reservoir reached its largest sizes. Additional random encoders continued to refine the structural pathway, although their marginal influence became less visible after integration with the joint neural readers, physics-only tree experts, and nonnegative prediction fusion.

The global protein–ligand branch produced a directionally consistent contribution throughout the reservoir-size study. For every evaluated value of $M$, retaining the global branch yielded a higher Pearson correlation and a lower RMSE than the corresponding global-branch ablation for both the complete model and the R-MLP-only path on both evaluation sets. The magnitude of this difference varied with reservoir size and dataset, ranging from modest changes at several head counts to more visible improvements in selected low-to-intermediate configurations. This pattern supports the global branch as a complementary structural view whose contribution persists across reservoir sizes, while its marginal value depends on the information already supplied by the ordinary random heads and the downstream expert pool.

Overall, reservoir expansion produced its clearest benefit in the structure-only R-MLP path, whereas the deterministic physicochemical representation and heterogeneous expert fusion re-duced the sensitivity of the complete model to the exact number of random encoders. The additional global protein–ligand branch supplied a smaller but consistently favorable contribution across all evaluated reservoir sizes.

**Tree-Based Physics Readers Complemented the Neural Expert Families**

To examine the contribution of different supervised readers, we performed a systematic regressor-family ablation under the fixed six-head RAVEN configuration. All variants retained the same structural reservoir, global protein–ligand branch, data partitions, and training protocol, while the composition of the supervised expert pool was varied. The single-family analysis evaluated R-MLP, J-MLP, Physics-MLP, ExtraTrees, XGBoost, and Random Forest as independent readers.[38,47,48] Three repeated models were used to characterize the performance variation within each family. The combination analysis further compared single-family ensembles, pairwise family combinations, the complete R-MLP/J-MLP/ExtraTrees ensemble, a single-instance three-family ensemble, and alternative replacements for the ExtraTrees physics reader. Table 3 summarizes the principal Pearson correlation and RMSE results.

**Table 3.** Regressor-family and physics-reader ablation under the fixed six-head RAVEN configuration. Panel A reports the mean $\pm$ standard deviation of three repeated single-reader models. Panel B reports prediction-level fusion results for different expert-family compositions. RMSE values are reported on the dimensionless pK scale. Bold values indicate the best result within each panel for the corresponding metric.

| Configuration | Test $r$ | Test RMSE | CASF $r$ | CASF RMSE |
|---|---|---|---|---|
| *Panel A. Single expert families* | | | | |
| R-MLP | 0.7687±0.0007 | 1.3158±0.0024 | 0.7921±0.0077 | 1.3261±0.0211 |
| J-MLP | 0.7735±0.0035 | 1.3042±0.0103 | 0.8046±0.0023 | 1.2907±0.0068 |
| Physics-MLP | 0.7395±0.0029 | 1.4156±0.0098 | 0.7726±0.0147 | 1.3870±0.0437 |
| ExtraTrees | **0.7781±0.0005** | 1.3085±0.0012 | **0.8196±0.0018** | 1.3071±0.0041 |
| XGBoost | 0.7772±0.0014 | **1.3010±0.0029** | 0.8178±0.0046 | **1.2651±0.0101** |
| Random Forest | 0.7578±0.0009 | 1.3639±0.0019 | 0.7977±0.0017 | 1.3547±0.0048 |
| *Panel B. Expert-family combinations and physics-reader replacements* | | | | |
| R-MLP ×3 + J-MLP ×3 + ExtraTrees ×3 | 0.7967 | 1.2463 | **0.8320** | 1.2403 |
| R-MLP ×3 | 0.7725 | 1.3052 | 0.7946 | 1.3190 |
| J-MLP ×3 | 0.7775 | 1.2929 | 0.8072 | 1.2831 |
| ExtraTrees ×3 | 0.7784 | 1.3078 | 0.8213 | 1.3064 |
| R-MLP ×3 + ExtraTrees ×3 | 0.7958 | 1.2507 | 0.8294 | 1.2530 |
| R-MLP ×3 + J-MLP ×3 | 0.7778 | 1.2912 | 0.8036 | 1.2937 |
| J-MLP ×3 + ExtraTrees ×3 | **0.7969** | 1.2465 | 0.8221 | 1.2498 |
| R-MLP ×1 + J-MLP ×1 + ExtraTrees ×1 | 0.7785 | 1.2978 | 0.8197 | 1.2728 |
| R-MLP ×3 + J-MLP ×3 + Physics-MLP ×3 | 0.7861 | 1.2687 | 0.8125 | 1.2676 |
| R-MLP ×3 + J-MLP ×3 + XGBoost ×3 | 0.7963 | **1.2441** | 0.8305 | **1.2323** |
| R-MLP ×3 + J-MLP ×3 + Random Forest ×3 | 0.7907 | 1.2601 | 0.8064 | 1.2626 |

Among the neural readers, J-MLP consistently provided a modest improvement over the structure-only R-MLP. On the PDBbind Test set, the mean Pearson correlation increased from 0.7687 for R-MLP to 0.7735 for J-MLP, while RMSE decreased from 1.3158 to 1.3042 pK. On CASF-2016, the corresponding Pearson correlations were 0.7921 and 0.8046, with RMSE values of 1.3261 and 1.2907 pK, respectively. These results show that incorporating the deterministic physicochemical fingerprint into the neural representation supplied additional predictive information, although the magnitude of the improvement remained limited.

A substantially different pattern emerged when the physicochemical fingerprint was read independently of the structural reservoir. Physics-MLP achieved Pearson correlations of 0.7395 on the PDBbind Test set and 0.7726 on CASF-2016. In comparison, ExtraTrees reached 0.7781 and 0.8196, while XGBoost reached 0.7772 and 0.8178, respectively. ExtraTrees also produced particularly small variation across repeated models in the correlation-based metrics, whereas XGBoost achieved the lowest RMSE and highest $R^2$ among the single-family readers on both evaluation sets. Random Forest remained competitive but was consistently below ExtraTrees and XGBoost in the principal correlation and error metrics. The contrast between Physics-MLP and the two stronger tree models indicates that the physicochemical fingerprint was more effectively exploited by ExtraTrees and XGBoost than by a conventional MLP reader.

The family-combination experiments reinforced this distinction. Combining R-MLP and J-MLP alone produced a Pearson correlation of 0.7778 and an RMSE of 1.2912 pK on the PDBbind Test set, with corresponding CASF-2016 values of 0.8036 and 1.2937 pK. Although this configuration already incorporated physicochemical information through J-MLP, both constituent readers remained MLP-based. Adding ExtraTrees produced a markedly stronger family combination. R-MLP plus ExtraTrees reached Pearson correlations of 0.7958 on the PDBbind Test set and 0.8294 on CASF-2016, while J-MLP plus ExtraTrees reached 0.7969 and 0.8221, respectively. Thus, the strongest pairwise configurations consistently contained the tree-based physics reader, supporting a complementary role for a non-neural regression family within the expert pool.

The complete R-MLP, J-MLP, and ExtraTrees ensemble provided the most balanced performance across the two evaluation sets. On the PDBbind Test set, the full three-family ensemble achieved a Pearson correlation of 0.7967 and an RMSE of 1.2463 pK. Its Pearson correlation was essentially tied with the J-MLP plus ExtraTrees combination at 0.7969, and their RMSE values differed by only 0.0002 pK. On CASF-2016, the complete ensemble achieved a Pearson correlation of 0.8320 and an RMSE of 1.2403 pK, outperforming all evaluated two-family combinations in correlation and overall error balance. The contribution of the third expert family was therefore not expressed as a large improvement on every individual metric, but the complete heterogeneous ensemble maintained the strongest overall profile across the two evaluation sets.

Expert multiplicity provided an additional source of improvement. Using one R-MLP, one J-MLP, and one ExtraTrees expert produced Pearson correlations of 0.7785 on the PDBbind Test set and 0.8197 on CASF-2016. When each family was represented by three independently fitted experts, the corresponding correlations increased to 0.7967 and 0.8320, while RMSE decreased from 1.2978 to 1.2463 pK on the PDBbind Test set and from 1.2728 to 1.2403 pK on CASF-2016. These results support repeated independent modeling within each expert family as a means of reducing dependence on a single supervised fit. Because other multiplicities were not systematically evaluated, this comparison supports the selected three-instance design without implying that three experts constitute a universally optimal family size.

Replacing the ExtraTrees pathway provided a further test of the importance of the physics-reader architecture. Substitution with Physics-MLP reduced Pearson correlation to 0.7861 on the PDBbind Test set and 0.8125 on CASF-2016, despite preserving the same physicochemical in-put pathway. Replacing ExtraTrees with Random Forest produced corresponding correlations of 0.7907 and 0.8064. XGBoost was the only alternative that closely matched the original heterogeneous ensemble, reaching Pearson correlations of 0.7963 and 0.8305. The XGBoost variant also achieved slightly lower RMSE values than the ExtraTrees configuration, with 1.2441 versus 1.2463 pK on the PDBbind Test set and 1.2323 versus 1.2403 pK on CASF-2016. ExtraTrees nevertheless retained slightly higher Pearson correlations in the complete ensemble and showed low repeated-model variation in its single-family correlation results. ExtraTrees and XGBoost therefore both emerged as effective tree-based physics readers, while Physics-MLP and Random Forest provided weaker alternatives under the evaluated configuration.

Taken together, the ablation results indicate that the physicochemical fingerprint was particularly well matched to tree-based readers such as ExtraTrees and XGBoost. The threshold partitioning and local feature interactions represented by tree ensembles may be especially suitable for the counts, extrema, histograms, distance bins, and gated interaction summaries contained in the physicochemical fingerprint. This model-form difference also introduced an inductive bias distinct from that of the MLP-based experts, providing complementary predictive behavior within the ensemble. Such behavior is consistent with broader observations that tree-based models can remain highly competitive with neural networks on medium-sized tabular datasets containing heterogeneous and irregular feature–response relationships.[49] Accordingly, the final RAVEN architecture combined multiple expert families to provide the regression stage with diverse inductive biases and to exploit their complementary strengths through prediction-level fusion. The use of multiple independently initialized experts within the same family further reduced dependence on any single supervised fit and contributed to the robustness of the ensemble. The overall benefit of the regressor architecture therefore arose from both heterogeneous diversity across expert families and repeated independent modeling within each family.

**Residual Complementarity Across Heterogeneous Expert Families**

The residual correlation matrices revealed a clear family-dependent structure among the nine experts in the final 32-head model. On the PDBbind Test set, residual correlations were extremely high within the R-MLP, J-MLP, and ExtraTrees families, with mean within-family correlations of approximately 0.991, 0.990, and 0.998, respectively. The two neural families were also strongly correlated with each other, with a mean R-MLP–J-MLP residual correlation of approximately 0.985. In contrast, the mean residual correlations between ExtraTrees and R-MLP and between ExtraTrees and J-MLP decreased to approximately 0.835 and 0.841, respectively. The same qualitative structure was retained on CASF-2016, indicating that the principal residual diversity arose across model families, particularly between the neural and tree-based experts, rather than from repeated random initialization within the same family.

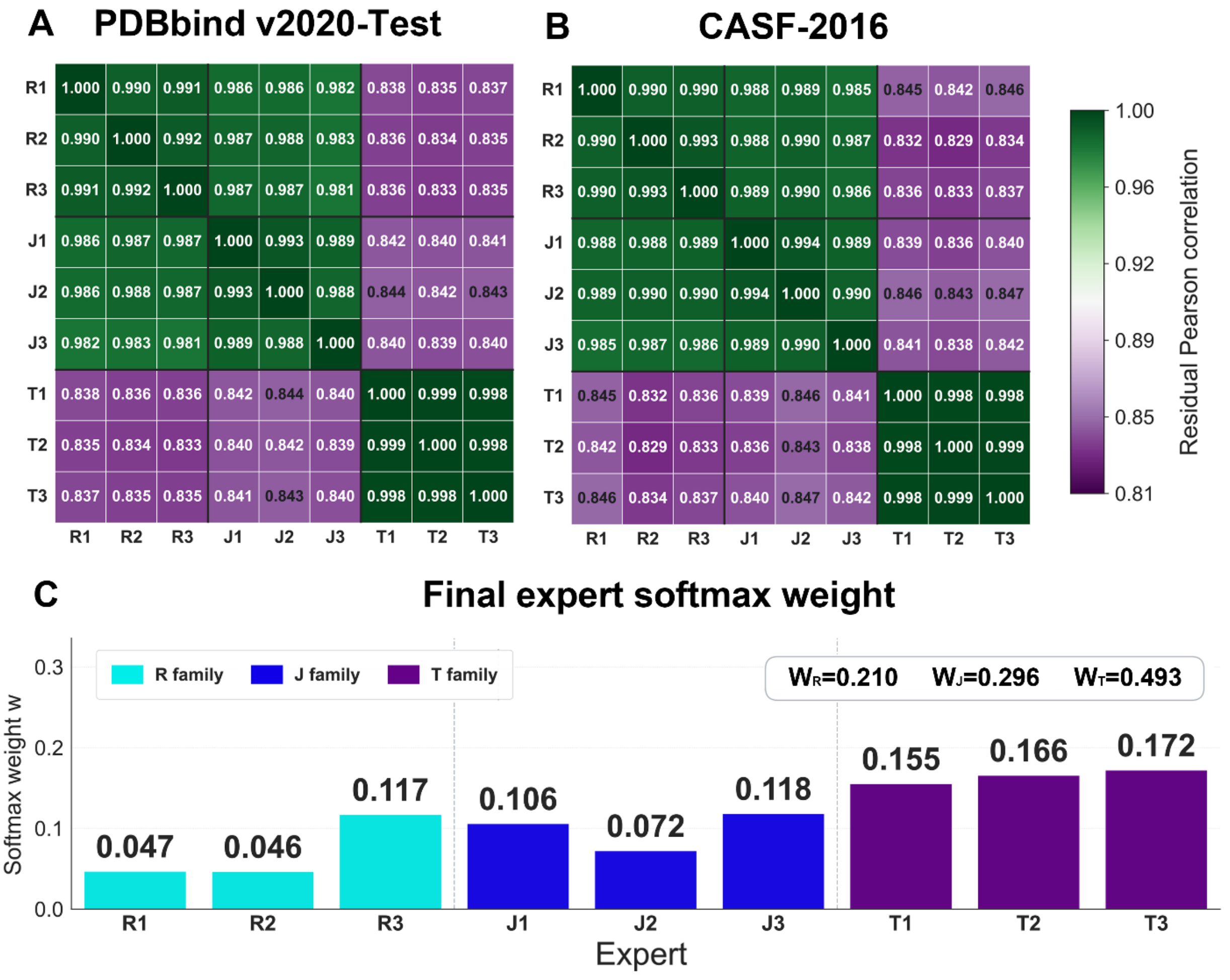

**Figure 4.** Residual dependence and learned ensemble weighting of the nine experts in the final 32-head RAVEN. (A) Pearson correlation matrix of expert residuals on the PDBbind Test set. (B) Corresponding residual correlation matrix on CASF-2016. (C) Final nonnegative softmax fusion weights learned exclusively on the Validation set. R1–R3, J1–J3, and T1–T3 denote the three R-MLP, J-MLP, and ExtraTrees experts, respectively.

The learned fusion weights were distributed across all nine experts rather than collapsing onto a single predictor, corresponding to an effective weight-based ensemble size of approximately $M_{\text{eff}} = 7.7$, where $M_{\text{eff}} = 1/\sum_k w_k^2$. At the family level, R-MLP, J-MLP, and ExtraTrees received total weights of 0.210, 0.296, and 0.493, respectively. The comparatively large ExtraTrees weight coincided with its substantially lower residual correlation with both neural families. Together with the preceding regressor ablation, these results support a structured form of expert complementarity: R-MLP and J-MLP produced highly similar error patterns, whereas the tree-based physics readers introduced a distinct residual structure that provided additional information to the final nonnegative fusion.

## Comparison with Representative Affinity Models

To place RAVEN in the context of established protein–ligand scoring approaches, we compared its performance with representative empirical, machine-learning, convolutional, graph-based, physics-informed, and interaction-fingerprint models on the CASF-2016 scoring-power benchmark.[43] As summarized in Table 4, classical AutoDock Vina and RF-Score-v3 provide historical reference points, while Pafnucy and $K_{\text{DEEP}}$ represent early three-dimensional convolutional approaches.[12,13,50,51] Because the published methods differ in training-set composition, PDBbind version, data filtering, feature construction, and model-selection procedures, the reported values are used as contextual references rather than as a strictly controlled leaderboard.

**Table 4.** Representative comparison of CASF-2016 scoring power. Pearson $r$ values are taken from the corresponding published studies. The methods differ in training data, preprocessing, feature construction, and model-selection protocols; therefore, the table is intended as a contextual comparison rather than a strictly controlled ranking.

| Method | Representation / learning strategy | Pearson $r$ |
|---|---|---|
| AutoDock Vina[43,50] | Empirical docking scoring function | 0.604 |
| RF-Score-v3[34,51] | Intermolecular distance counts and Vina terms with Random Forest | 0.803 |
| Pafnucy[13] | Three-dimensional voxel representation with CNN | 0.780 |
| $K_{\text{DEEP}}$[12] | Three-dimensional voxel representation with CNN | 0.820 |
| PLEC[34] | Protein–ligand extended-connectivity fingerprint | 0.817 |
| AEScore[37] | Atomic-environment representation with neural regression | 0.830 |
| PIGNet ensemble[19] | Physics-informed graph neural network | 0.761 |
| InteractionGraphNet[17] | Learned intra- and intermolecular graph interactions | 0.837 |
| **RAVEN** | Frozen random graph reservoir, explicit physics, and heterogeneous expert fusion | **0.8446** |
| OnionNet-2[36] | Multi-shell residue–atom contacts with CNN | 0.864 |
| ECIF6::LD-GBT[15] | Extended-connectivity interaction features and ligand descriptors with GBT | 0.866 |
| Multi-shelled ECIF[52] | Distance-resolved interaction fingerprint with GBT | 0.877 |

Among learned structure-based methods, AEScore and InteractionGraphNet reported Pearson correlations of 0.830 and 0.837, respectively, whereas RAVEN reached 0.8446 despite keeping all graph-reservoir encoders frozen during supervised regression.[17,37] Interaction-fingerprint and multi-shell approaches remained particularly strong: PLEC achieved 0.817, OnionNet-2 reached 0.864, ECIF6::LD-GBT reached 0.866, and Multi-shelled ECIF reported 0.877.[15,34,36,52] These results place RAVEN within the competitive performance range of established three-dimensional and graph-based affinity models, while several specialized interaction-descriptor approaches retain higher reported CASF-2016 correlations.

The comparison therefore does not establish RAVEN as a universal state-of-the-art scoring function, but shows that competitive CASF-2016 scoring performance can be obtained without end-to-end optimization of the graph representation. Together with the preceding ablation results, this observation supports the use of complementary representation and regression strategies: fro-zen randomized graph views provide structural diversity, explicit physicochemical descriptors retain interaction information, and heterogeneous expert families introduce distinct supervised inductive biases. RAVEN thus provides an alternative affinity-prediction architecture in which competitive performance emerges from the combination of complementary representations and readers rather than from jointly training a single increasingly complex representation model.

## Discussion

The present experiments support the central design hypothesis of RAVEN: competitive protein–ligand affinity regression can be obtained by combining multiple independently sampled frozen graph representations with explicit physicochemical descriptors and heterogeneous supervised readers. The reservoir-size experiments showed that increasing the number of independent frozen MPNN realizations generally improved correlation and reduced prediction error, although the gains became smaller and locally nonmonotonic at larger reservoir sizes. The regressor-family ablations further showed that combining predictors with different functional forms provided greater benefit than simply combining closely related neural readers. Together, these observations support two closely connected ideas underlying the architecture: replication of independent frozen graph encoders expands the structural representation available to the supervised stage, while heterogeneous representation and regression families provide complementary inductive biases that reduce dependence on any single modeling pathway. RAVEN therefore derives its predictive performance from a division of representational and functional roles rather than from increasingly intensive optimization of one end-to-end neural model.

Although the parameters governing message generation and node updating in each reservoir encoder are randomly initialized, the resulting representation is not independent of molecular structure. Message propagation remains constrained by the atomistic graph topology, relation types, covalent connectivity, cross-interface contacts, atomic attributes, and geometric edge information. Each frozen MPNN therefore implements a randomly parameterized but structure-conditioned nonlinear transformation of the same protein–ligand complex. This interpretation is closely related to random-feature learning, in which fixed randomized nonlinear mappings can provide useful supervised feature spaces and increasing numbers of independent random features improve the approximation of an underlying induced kernel in expectation.[27] Related graph-reservoir and graph-random-feature studies extend the same general principle to structured graph inputs.[25,26] In RAVEN, the Monte Carlo unit is an entire independently initialized MPNN rather than an individual hidden coordinate, so increasing the head count samples additional complete graph mappings from the initialization-induced encoder distribution. The observed head-count trend is consistent with this interpretation: additional random encoders improved the readability of the structure-only representation on average, whereas diminishing returns and local reversals remained possible because the supervised readout dimension increased simultaneously under a fixed amount of training data.[33] Frozen random encoders should therefore be viewed as task-independent nonlinear structural projections rather than as unstructured random noise.

The deterministic physicochemical fingerprint was introduced to complement, rather than replace or correct, the randomized structural reservoir. Its construction deliberately relies on comparatively simple chemical and geometric relationships instead of a trainable energy decomposition. Molecular composition, intramolecular topology, interface-distance distributions, contact geometry, and channel-resolved interaction events are concatenated into a fixed representation whose coordinates retain the same chemical meanings across samples and random seeds. This fixed semantic structure is fundamentally different from the initialization-dependent coordinates produced by the frozen MPNNs. The two pathways therefore expose the supervised regressors to complementary forms of information: the reservoir provides multiple nonlinear views of atomistic structure, whereas the physicochemical branch provides explicitly defined and reproducible summaries of molecular and intermolecular properties. The strong performance of interaction-fingerprint and multiscale-contact approaches such as ECIF, PLEC, and OnionNet-2 further demonstrates that carefully constructed fixed interaction representations remain highly informative for protein–ligand affinity prediction.[15,34,36] In the present experiments, the physicochemical pathway likewise supplied useful information beyond the frozen structural representation, supporting the decision to retain explicit chemical semantics alongside randomized graph features rather than requiring either representation family to recover the complete affinity relationship alone.

The regressor ablations further indicate that useful descriptors and suitable supervised readers should be considered jointly. When the same physicochemical fingerprint was processed by different model families, ExtraTrees and XGBoost consistently produced stronger correlation-based performance than the conventional Physics-MLP, indicating that the representation was particularly compatible with tree-based regression. The fingerprint contains numerous counts, extrema, distance bins, histograms, gated events, and local interaction summaries, for which recursive threshold partitioning and local feature conjunctions provide a plausible functional match.[38,47,49] More importantly, the tree-based readers introduced a supervised inductive bias distinct from that of the two MLP families. This functional diversity was directly reflected in the residual analysis: errors were highly correlated among repeated members of the same family and between R-MLP and J-MLP, whereas the ExtraTrees residuals showed substantially lower correlation with both neural families. The learned nonnegative fusion consequently distributed weight across multiple experts while assigning a substantial fraction of the total weight to the tree-based family. These observations provide direct experimental support for the complementary-design principle of RAVEN: the most useful diversity arose from combining different representation and model families rather than from simply replicating nearly redundant predictors. Consistent with this interpretation, the final frozen-encoder model reached a CASF-2016 scoring accuracy within the competitive range of established three-dimensional and graph-based affinity models, although several specialized interaction-fingerprint approaches retained higher reported correlations.

Several aspects of the present study define the scope within which these conclusions should be interpreted. The reported experiments evaluate affinity scoring from experimentally resolved protein–ligand complex structures and therefore do not establish performance for pose ranking, docking-power evaluation, or virtual screening. The similarity-isolated partition prevents cross-subset pairs satisfying the predefined ligand-, protein-, and complex-level conflict criteria, but it does not imply complete novelty under every possible definition of molecular or structural similarity. Freezing the graph encoders guarantees that affinity gradients cannot modify their structural projections, but the present experiments were designed to establish the predictive viability of frozen reservoirs rather than to demonstrate universal superiority over otherwise identical end-to-end trained encoders or to directly quantify suppression of dataset-induced shortcuts. The deterministic physicochemical representation is likewise intentionally simplified: its interaction channels summarize chemically motivated geometric patterns and should not be interpreted as an additive decomposition of binding free energy. Finally, the observed reservoir-size curves showed diminishing returns at moderate-to-large head counts, while prediction errors remained more pronounced in sparsely populated regions at the extremes of the affinity distribution. These observations indicate that increasing representation capacity alone does not remove limitations associated with finite training data and uneven coverage of the target distribution.

These findings suggest several natural extensions of the present framework. Evaluation on temporally separated, prospective, or more aggressively similarity-controlled external datasets would provide a stronger test of the transferability of frozen graph reservoirs beyond the current PDBbind-derived setting. Pose-ranking and virtual-screening experiments could further determine whether the complementary structural and physicochemical pathways remain useful when the input geometry itself becomes uncertain. Additional physicochemical channels, alternative graph-propagation architectures, and more systematic control of reservoir size could also be explored without changing the basic separation between frozen structural encoding and supervised readout. A direct comparison between frozen and end-to-end trained versions of the same encoder family would further help distinguish the effects of task-independent random projection from those of supervised representation adaptation. More broadly, RAVEN provides a modular setting in which representation diversity and model-family diversity can be varied independently, offering a practical framework for studying how complementary inductive biases can be combined in protein–ligand affinity prediction.

## Conclusion

RAVEN presents a protein–ligand binding-affinity prediction framework that combines independently initialized and fully frozen atomistic graph encoders with an explicit physicochemical interaction fingerprint and heterogeneous supervised readers. The results show that frozen random MPNN representations retain useful structure-dependent information and that increasing the number of independent encoder realizations generally improves the predictive utility of the structural reservoir. The physicochemical pathway provides complementary fixed chemical semantics, while tree-based readers such as ExtraTrees and XGBoost effectively exploit these descriptors and introduce predictive behavior distinct from that of the neural expert families. The resulting heterogeneous fusion therefore benefits from complementarity across both representation types and supervised inductive biases rather than relying on a single optimized representation pathway.

The final RAVEN model achieved a Pearson correlation of 0.7995 on the similarity-isolated PDBbind 2020R1 Test set and 0.8446 on the protected CASF-2016 subset while keeping all graph encoders fixed throughout supervised learning. These findings indicate that useful protein–ligand representations do not require every component to be optimized jointly: randomized structural projections, explicit physicochemical information, and heterogeneous regression models can instead be combined to exploit complementary strengths. Future work can extend this framework to external and temporally separated datasets, alternative structure-based prediction tasks, and broader physicochemical representations to further examine the generality of frozen graph reservoirs for molecular prediction.

## Acknowledgements

This work was supported by the National Natural Science Foundation of China (Grant 22373116, and 62406343), and the Natural Science Foundation of Jiangsu Province(Grants No BK20252072). The Supercomputer Center of China Pharmaceutical University is acknowledged for providing computer resources. ChatGPT (OpenAI) was used for assistance with programming, code debugging, and language refinement during manuscript preparation. All scientific analyses, interpretation of results, and final manuscript content were reviewed and determined by the authors.

## References

(1) Gilson, M. K.; Zhou, H.-X. Calculation of Protein–Ligand Binding Affinities. *Annual Review of Biophysics and Biomolecular Structure* **2007**, *36*, 21-42.

(2) Olsson, T. S. G.; Williams, M. A.; Pitt, W. R.; Ladbury, J. E. The Thermodynamics of Protein–Ligand Interaction and Solvation: Insights for Ligand Design. *Journal of Molecular Biology* **2008**, *384*, 1002-1017.

(3) Reynolds, C. H.; Holloway, M. K. Thermodynamics of Ligand Binding and Efficiency. *ACS Medicinal Chemistry Letters* **2011**, *2*, 433-437.

(4) Wang, R.; Fang, X.; Lu, Y.; Yang, C.-Y.; Wang, S. The PDBbind Database: Methodologies and Updates. *Journal of Medicinal Chemistry* **2005**, *48*, 4111-4119.

(5) Meli, R.; Morris, G. M.; Biggin, P. C. Scoring Functions for Protein–Ligand Binding Affinity Prediction Using Structure-Based Deep Learning: A Review. *Frontiers in Bioinformatics* **2022**, *2*, 885983.

(6) Gorantla, R.; Kubincová, A.; Weiße, A. Y.; Mey, A. S. J. S. From Proteins to Ligands: Decoding Deep Learning Methods for Binding Affinity Prediction. *Journal of Chemical Information and Modeling* **2024**, *64*, 2496-2507.

(7) Kalliokoski, T.; Kramer, C.; Vulpetti, A.; Gedeck, P. Comparability of Mixed IC50 Data–A Statistical Analysis. *PLoS ONE* **2013**, *8*, e61007.

(8) Li, J.; Guan, X.; Zhang, O.; Sun, K.; Wang, Y.; Bagni, D.; Head-Gordon, T. Leak Proof PDBBind: A Reorganized Dataset of Protein–Ligand Complexes for More Generalizable Binding Affinity Prediction. *Journal of Physical Chemistry B* **2026**, *130*, 730-740.

(9) Francoeur, P. G.; Masuda, T.; Sunseri, J.; Jia, A.; Iovanisci, R. B.; Snyder, I.; Koes, D. R. Three-Dimensional Convolutional Neural Networks and a Cross-Docked Data Set for Structure-Based Drug Design. *Journal of Chemical Information and Modeling* **2020**, *60*, 4200-4215.

(10) Su, M.; Yang, Q.; Du, Y.; Feng, G.; Liu, Z.; Li, Y.; Wang, R. Comparative Assessment of Scoring Functions: The CASF-2016 Update. *Journal of Chemical Information and Modeling* **2019**, *59*, 895-913.

(11) Kanakala, G. C.; Aggarwal, R.; Nayar, D.; Priyakumar, U. D. Latent Biases in Machine Learning Models for Predicting Binding Affinities Using Popular Data Sets. *ACS Omega* **2023**, *8*, 2389-2397.

(12) Jiménez, J.; Škalič, M.; Martínez-Rosell, G.; De Fabritiis, G. KDEEP: Protein–Ligand Absolute Binding Affinity Prediction via 3D-Convolutional Neural Networks. *Journal of Chemical Information and Modeling* **2018**, *58*, 287-296.

(13) Stepniewska-Dziubinska, M. M.; Zielenkiewicz, P.; Siedlecki, P. Development and Evaluation of a Deep Learning Model for Protein–Ligand Binding Affinity Prediction. *Bioinformatics* **2018**, *34*, 3666-3674.

(14) Ballester, P. J.; Mitchell, J. B. O. A Machine Learning Approach to Predicting Protein–Ligand Binding Affinity with Applications to Molecular Docking. *Bioinformatics* **2010**, *26*, 1169-1175.

(15) Sánchez-Cruz, N.; Medina-Franco, J. L.; Mestres, J.; Barril, X. Extended Connectivity Interaction Features: Improving Binding Affinity Prediction through Chemical Description. *Bioinformatics* **2021**, *37*, 1376-1382.

(16) Feinberg, E. N.; Sur, D.; Wu, Z.; Husic, B. E.; Mai, H.; Li, Y.; Sun, S.; Yang, J.; Ramsundar, B.; Pande, V. S. PotentialNet for Molecular Property Prediction. *ACS Central Science* **2018**, *4*, 1520-1530.

(17) Jiang, D.; Hsieh, C.-Y.; Wu, Z.; Kang, Y.; Wang, J.; Wang, E.; Liao, B.; Shen, C.; Xu, L.; Wu, J.; Cao, D.; Hou, T. InteractionGraphNet: A Novel and Efficient Deep Graph Representation Learning Framework for Accurate Protein–Ligand Interaction Predictions. *Journal of Medicinal Chemistry* **2021**, *64*, 18209-18232.

(18) Karlov, D. S.; Sosnin, S.; Fedorov, M. V.; Popov, P. graphDelta: MPNN Scoring Function for the Affinity Prediction of Protein–Ligand Complexes. *ACS Omega* **2020**, *5*, 5150-5159.

(19) Moon, S.; Zhung, W.; Yang, S.; Lim, J.; Kim, W. Y. PIGNet: A Physics-Informed Deep Learning Model toward Generalized Drug–Target Interaction Predictions. *Chemical Science* **2022**, *13*, 3661-3673.

(20) Volkov, M.; Turk, J.-A.; Drizard, N.; Martin, N.; Hoffmann, B.; Gaston-Mathé, Y.; Rognan, D. On the Frustration to Predict Binding Affinities from Protein–Ligand Structures with Deep Neural Networks. *Journal of Medicinal Chemistry* **2022**, *65*, 7946-7958.

(21) Mastropietro, A.; Pasculli, G.; Bajorath, J. Learning Characteristics of Graph Neural Networks Predicting Protein–Ligand Affinities. *Nature Machine Intelligence* **2023**, *5*, 1427-1436.

(22) Wallach, I.; Heifets, A. Most Ligand-Based Classification Benchmarks Reward Memorization Rather than Generalization. *Journal of Chemical Information and Modeling* **2018**, *58*, 916-932.

(23) Joeres, R.; Blumenthal, D. B.; Kalinina, O. V. Data Splitting to Avoid Information Leakage with DataSAIL. *Nature Communications* **2025**, *16*, 3337.

(24) Graber, D.; Stockinger, P.; Meyer, F.; Mishra, S.; Horn, C.; Buller, R. Resolving Data Bias Improves Generalization in Binding Affinity Prediction. *Nature Machine Intelligence* **2025**, *7*, 1713-1725.

(25) Gallicchio, C.; Micheli, A. *In 2010 international joint conference on neural networks*, IEEE: 2010, pp 2159-2166.

(26) Zambon, D.; Alippi, C.; Livi, L. *In Proceedings of the 37th international conference on machine learning*, PMLR: 2020; Vol. 119, pp 10968-10977.

(27) Rahimi, A.; Recht, B. *In Advances in neural information processing systems*, Curran Associates, Inc.: 2007; Vol. 20, pp 1177-1184.

(28) Dong, J.; Ohana, R.; Rafayelyan, M.; Krzakala, F. *In Advances in neural information processing systems*, 2020; Vol. 33.

(29) Lee, H.-J.; Emani, P. S.; Gerstein, M. B. Improved Prediction of Ligand–Protein Binding Affinities by Metamodeling. *Journal of Chemical Information and Modeling* **2024**, *64*, 8684-8704.

(30) Mohamed Abdul Cader, J.; Newton, M. A. H.; Rahman, J.; Mohamed Abdul Cader, A. J.; Sattar, A. Ensembling Methods for Protein–Ligand Binding Affinity Prediction. *Scientific Reports* **2024**, *14*, 24447.

(31) Gilmer, J.; Schoenholz, S. S.; Riley, P. F.; Vinyals, O.; Dahl, G. E. *In Proceedings of the 34th international conference on machine learning*, PMLR: 2017; Vol. 70, pp 1263-1272.

(32) Wang, M.; Zheng, D.; Ye, Z.; Gan, Q.; Li, M.; Song, X.; Zhou, J.; Ma, C.; Yu, L.; Gai, Y.; Xiao, T.; He, T.; Karypis, G.; Li, J.; Zhang, Z. Deep Graph Library: A Graph-Centric, Highly-Performant Package for Graph Neural Networks. *arXiv preprint arXiv:1909.01315* **2019**.

(33) Meng, X.; Yao, J.; Cao, Y. Multiple Descent in the Multiple Random Feature Model. *Journal of Machine Learning Research* **2024**, *25*, 1-49.

(34) Wójcikowski, M.; Kukiełka, M.; Stępniewska-Dziubinska, M. M.; Siedlecki, P. Development of a Protein–Ligand Extended Connectivity (PLEC) Fingerprint and Its Application for Binding Affinity Predictions. *Bioinformatics* **2019**, *35*, 1334-1341.

(35) Zheng, L.; Fan, J.; Mu, Y. OnionNet: A Multiple-Layer Intermolecular-Contact-Based Convolutional Neural Network for Protein–Ligand Binding Affinity Prediction. *ACS Omega* **2019**, *4*, 15956-15965.

(36) Wang, Z.; Zheng, L.; Liu, Y.; Qu, Y.; Li, Y.-Q.; Zhao, M.; Mu, Y.; Li, W. OnionNet-2: A Convolutional Neural Network Model for Predicting Protein–Ligand Binding Affinity Based on Residue–Atom Contacting Shells. *Frontiers in Chemistry* **2021**, *9*, 753002.

(37) Meli, R.; Anighoro, A.; Bodkin, M. J.; Morris, G. M.; Biggin, P. C. Learning Protein–Ligand Binding Affinity with Atomic Environment Vectors. *Journal of Cheminformatics* **2021**, *13*, 59.

(38) Geurts, P.; Ernst, D.; Wehenkel, L. Extremely Randomized Trees. *Machine Learning* **2006**, *63*, 3-42.

(39) Wolpert, D. H. Stacked Generalization. *Neural Networks* **1992**, *5*, 241-259.
(40) Liu, Z.; Li, Y.; Han, L.; Li, J.; Liu, J.; Zhao, Z.; Nie, W.; Liu, Y.; Wang, R. PDB-Wide Collection of Binding Data: Current Status of the PDBbind Database. *Bioinformatics* **2015**, *31*, 405-412.
(41) Graber, D.; Stockinger, P.; Meyer, F.; Mishra, S.; Horn, C.; Buller, R. Resolving Data Bias Improves Generalization in Binding Affinity Prediction. *Nature Machine Intelligence* **2025**, *7*, 1713-1725.
(42) Stockinger, P. GEMS: Resolving Data Bias Improves Generalization in Binding Affinity Prediction, version 3, Zenodo: 2025.
(43) Su, M.; Yang, Q.; Du, Y.; Feng, G.; Liu, Z.; Li, Y.; Wang, R. Comparative Assessment of Scoring Functions: The CASF-2016 Update. *Journal of Chemical Information and Modeling* **2019**, *59*, 895-913.
(44) Cawley, G. C.; Talbot, N. L. C. On Over-Fitting in Model Selection and Subsequent Selection Bias in Performance Evaluation. *Journal of Machine Learning Research* **2010**, *11*, 2079-2107.
(45) Paszke, A. et al. *In Advances in neural information processing systems*, 2019; Vol. 32, pp 8024-8035.
(46) Pedregosa, F. et al. Scikit-learn: Machine Learning in Python. *Journal of Machine Learning Research* **2011**, *12*, 2825-2830.
(47) Chen, T.; Guestrin, C. *In Proceedings of the 22nd ACM SIGKDD international conference on knowledge discovery and data mining*, 2016, pp 785-794.
(48) Breiman, L. Random Forests. *Machine Learning* **2001**, *45*, 5-32.
(49) Grinsztajn, L.; Oyallon, E.; Varoquaux, G. *In Advances in neural information processing systems*, 2022; Vol. 35.
(50) Trott, O.; Olson, A. J. AutoDock Vina: Improving the Speed and Accuracy of Docking with a New Scoring Function, Efficient Optimization, and Multithreading. *Journal of Computational Chemistry* **2010**, *31*, 455-461.
(51) Li, H.; Leung, K.-S.; Wong, M.-H.; Ballester, P. J. Low-Quality Structural and Interaction Data Improves Binding Affinity Prediction via Random Forest. *Molecules* **2015**, *20*, 10947-10962.
(52) Shiota, K.; Akutsu, T. Multi-shelled ECIF: Improved Extended Connectivity Interaction Features for Accurate Binding Affinity Prediction. *Bioinformatics Advances* **2023**, *3*, vbad155.
(53) Hoeffding, W. Probability Inequalities for Sums of Bounded Random Variables. *Journal of the American Statistical Association* **1963**, *58*, 13-30.
(54) Lee, J.; Bahri, Y.; Novak, R.; Schoenholz, S. S.; Pennington, J.; Sohl-Dickstein, J. *In International conference on learning representations*, 2018.

# Appendix: Justification of Viewpoint and Extended Analysis of the Random-Feature Reservoir

## A1. Random-Feature Construction and Initialization-Induced Kernel

Let $G$ denote a protein–ligand heterograph and let $\omega \sim P_\Omega$ denote one complete random realization of an ordinary frozen graph encoder. Here, $\omega$ comprises the complete parameterized mapping, including the node and edge encoders, relation-specific message functions, recurrent update modules, normalization parameters, and cross-interface pair network. The random unit considered in the following analysis is therefore an entire independently initialized encoder rather than an individual hidden coordinate.

For one realization $\omega$, the corresponding graph-level feature block is

$$\boldsymbol{\phi}_\omega(G) = [\mathbf{p}_\omega(G); \mathbf{i}_\omega(G); \mathbf{l}_\omega(G)] \in \mathbb{R}^{1152}. \tag{14}$$

For $M$ ordinary encoder realizations,

$$\omega_1, \dots, \omega_M \overset{\text{i.i.d.}}{\sim} P_\Omega, \tag{15}$$

the ordinary-head reservoir is obtained by block concatenation,

$$\mathbf{C}_M(G) = \left[\boldsymbol{\phi}_{\omega_1}(G); \cdots; \boldsymbol{\phi}_{\omega_M}(G)\right] \in \mathbb{R}^{1152M}. \tag{16}$$

The separately initialized global branch used in the complete RAVEN representation is excluded from the present Monte Carlo analysis because it employs a different graph-level readout and is not treated as an identically distributed ordinary-head realization.

For two heterographs $G$ and $G'$, define the scale-normalized similarity induced by the concatenated ordinary reservoir as

$$\widehat{K}_M(G, G') = \frac{1}{M}\mathbf{C}_M(G)^\top \mathbf{C}_M(G') = \frac{1}{M}\sum_{m=1}^{M} \boldsymbol{\phi}_{\omega_m}(G)^\top \boldsymbol{\phi}_{\omega_m}(G'). \tag{17}$$

Thus, although the implemented representation retains all encoder outputs by concatenation, the scale-normalized inner product induced by that representation is the arithmetic mean of the similarity contributions from the independent encoder realizations.

For one realization, define

$$X_\omega(G, G') = \boldsymbol{\phi}_\omega(G)^\top \boldsymbol{\phi}_\omega(G'), \tag{18}$$

and its initialization-averaged expectation

$$K_\Omega(G, G') = \mathbb{E}_{\omega \sim P_\Omega}[X_\omega(G, G')]. \tag{19}$$

Whenever the required expectations are finite, $K_\Omega$ is positive semidefinite. For any finite collection of graphs $G_1, \dots, G_N$ and real coefficients $c_1, \dots, c_N$,

$$\begin{aligned} \sum_{a=1}^{N}\sum_{b=1}^{N} c_a c_b K_\Omega(G_a, G_b) &= \mathbb{E}_\omega\left[\sum_{a=1}^{N}\sum_{b=1}^{N} c_a c_b \boldsymbol{\phi}_\omega(G_a)^\top \boldsymbol{\phi}_\omega(G_b)\right] \\ &= \mathbb{E}_\omega\left[\left\|\sum_{a=1}^{N} c_a \boldsymbol{\phi}_\omega(G_a)\right\|_2^2\right] \geq 0. \end{aligned} \tag{20}$$

Accordingly, $K_\Omega$ defines an initialization-induced positive-semidefinite graph kernel. No claim is made that the finite heterogeneous MPNN family separates all non-isomorphic graphs.

## A2. Justification of our viewpoint

**Our viewpoint.** Assume that the complete ordinary-encoder realizations $\omega_1, \dots, \omega_M$ are independent and identically distributed and that, for fixed $G$ and $G'$,

$$\mathrm{Var}[X_\omega(G, G')] = \sigma_{G,G'}^2 < \infty.$$

Then

$$\mathbb{E}\left[\widehat{K}_M(G, G')\right] = K_\Omega(G, G'), \qquad \mathrm{Var}\left[\widehat{K}_M(G, G')\right] = \frac{\sigma_{G,G'}^2}{M}, \qquad \widehat{K}_M(G, G') - K_\Omega(G, G') = O_p\left(M^{-1/2}\right).$$

(21)

*Proof.* From Equation 17,

$$\widehat{K}_M(G,G') = \frac{1}{M}\sum_{m=1}^{M} X_{\omega_m}(G,G'). \tag{22}$$

By linearity of expectation and identical distribution of the complete encoder realizations,

$$\begin{aligned}\mathbb{E}\big[\widehat{K}_M(G,G')\big] &= \frac{1}{M}\sum_{m=1}^{M} \mathbb{E}\left[X_{\omega_m}(G,G')\right] \\ &= \frac{1}{M}\sum_{m=1}^{M} K_\Omega(G,G') = K_\Omega(G,G').\end{aligned} \tag{23}$$

Hence, the empirical similarity is an unbiased estimator of the initialization-averaged kernel.

For the variance,

$$\begin{aligned}\mathrm{Var}\big[\widehat{K}_M(G,G')\big] &= \frac{1}{M^2}\mathrm{Var}\big[\textstyle\sum_{m=1}^{M} X_{\omega_m}(G,G')\big] \\ &= \frac{1}{M^2}\big[\textstyle\sum_{m=1}^{M}\mathrm{Var}\left(X_{\omega_m}\right) + 2\sum_{m<n}\mathrm{Cov}\left(X_{\omega_m},X_{\omega_n}\right)\big].\end{aligned} \tag{24}$$

Because the complete encoder realizations are independent,

$$\mathrm{Cov}\big(X_{\omega_m},X_{\omega_n}\big) = 0, \qquad m \neq n.$$

Identical distribution further gives

$$\mathrm{Var}\big[X_{\omega_m}(G,G')\big] = \sigma^2_{G,G'}.$$

Therefore,

$$\mathrm{Var}\big[\widehat{K}_M(G,G')\big] = \frac{1}{M^2}M\sigma^2_{G,G'} = \frac{\sigma^2_{G,G'}}{M}. \tag{25}$$

To establish the stochastic order, define

$$Z_M = \sqrt{M}\big[\widehat{K}_M(G,G') - K_\Omega(G,G')\big]. \tag{26}$$

The preceding results imply

$$\mathbb{E}[Z_M] = 0, \qquad \mathrm{Var}(Z_M) = M\mathrm{Var}\big(\widehat{K}_M\big) = \sigma^2_{G,G'}.$$

For any $C > 0$, Chebyshev's inequality gives

$$\Pr(|Z_M| \geq C) \leq \frac{\sigma^2_{G,G'}}{C^2}. \tag{27}$$

For any $\delta > 0$, choosing $C > \sigma_{G,G'}/\sqrt{\delta}$ therefore yields

$$\sup_M \Pr(|Z_M| > C) < \delta.$$

Hence $Z_M = O_p(1)$, and consequently

$$\widehat{K}_M(G,G') - K_\Omega(G,G') = O_p\big(M^{-1/2}\big). \tag{28}$$

This completes the proof. ▫

The independence assumption in the viewpoint concerns the complete encoder realizations $\omega_m$. No independence assumption is made for the 1152 coordinates within an individual feature block $\boldsymbol{\phi}_{\omega_m}(G)$, which share the same multilayer parameter realization.

## A3. Finite-Sample Concentration Bounds

The finite-variance assumption immediately gives a nonasymptotic concentration bound. By Chebyshev's inequality and Equation 25,

$$\Pr\big(\big|\widehat{K}_M(G,G') - K_\Omega(G,G')\big| \geq \varepsilon\big) \leq \frac{\sigma^2_{G,G'}}{M\varepsilon^2}. \tag{29}$$

This result requires only a finite second moment.

Under the additional analytical condition

$$|X_\omega(G,G')| \le B \quad \text{almost surely,} \tag{30}$$

each similarity contribution lies in the interval $[-B,B]$. Hoeffding's inequality then gives[27,53]

$$\begin{aligned}\Pr\left(\left|\hat{K}_M - K_\Omega\right| \ge \varepsilon\right) &\le 2\exp\left[-\frac{2M^2\varepsilon^2}{\sum_{m=1}^{M}(2B)^2}\right] \\ &= 2\exp\left(-\frac{M\varepsilon^2}{2B^2}\right).\end{aligned} \tag{31}$$

The boundedness assumption is required only for this exponential bound and is not required for our viewpoint.

## A4. Corresponding Concentration of Reservoir-Induced Distances

The same block-sampling argument applies to the scale-normalized squared Euclidean distance induced by the concatenated ordinary reservoir. Define

$$\begin{aligned}\hat{D}_M^2(G,G') &= \frac{1}{M}\|\mathbf{C}_M(G) - \mathbf{C}_M(G')\|_2^2 \\ &= \frac{1}{M}\sum_{m=1}^{M}\left\|\boldsymbol{\phi}_{\omega_m}(G) - \boldsymbol{\phi}_{\omega_m}(G')\right\|_2^2.\end{aligned} \tag{32}$$

For one encoder realization, define

$$Y_\omega(G,G') = \|\boldsymbol{\phi}_\omega(G) - \boldsymbol{\phi}_\omega(G')\|_2^2 \tag{33}$$

and

$$D_\Omega^2(G,G') = \mathbb{E}_\omega[Y_\omega(G,G')]. \tag{34}$$

If

$$\mathrm{Var}[Y_\omega(G,G')] = \tau_{G,G'}^2 < \infty,$$

then independence of the complete encoder realizations gives

$$\mathbb{E}\left[\hat{D}_M^2(G,G')\right] = D_\Omega^2(G,G'), \qquad \mathrm{Var}\left[\hat{D}_M^2(G,G')\right] = \frac{\tau_{G,G'}^2}{M}, \tag{35}$$

and, by the same Chebyshev argument,

$$\hat{D}_M^2(G,G') - D_\Omega^2(G,G') = O_p\left(M^{-1/2}\right). \tag{36}$$

The quantity $D_\Omega^2$ is an initialization-averaged feature-space distance. No claim is made that it defines a complete metric capable of separating all non-isomorphic graphs.

## A5. Expected Diversity between Independent Random Graph Views

Independent complete encoder realizations also provide a direct representation-level measure of initialization diversity. For fixed $G$, let

$$\mathbf{Z} = \boldsymbol{\phi}_\omega(G), \qquad \mathbf{Z}' = \boldsymbol{\phi}_{\omega'}(G),$$

where $\omega, \omega' \overset{\text{i.i.d.}}{\sim} P_\Omega$. Then

$$\begin{aligned}\mathbb{E}\left[\|\mathbf{Z} - \mathbf{Z}'\|_2^2\right] &= \mathbb{E}[\mathbf{Z}^\top\mathbf{Z} + \mathbf{Z}'^\top\mathbf{Z}' - 2\mathbf{Z}^\top\mathbf{Z}'] \\ &= 2\mathbb{E}[\mathbf{Z}^\top\mathbf{Z}] - 2\mathbb{E}[\mathbf{Z}]^\top\mathbb{E}[\mathbf{Z}].\end{aligned} \tag{37}$$

Using

$$\mathrm{tr}[\mathrm{Cov}(\mathbf{Z})] = \mathbb{E}[\mathbf{Z}^\top\mathbf{Z}] - \mathbb{E}[\mathbf{Z}]^\top\mathbb{E}[\mathbf{Z}],$$

it follows that

$$\mathbb{E}\left[\|\boldsymbol{\phi}_\omega(G) - \boldsymbol{\phi}_{\omega'}(G)\|_2^2\right] = 2\mathrm{tr}\left[\mathrm{Cov}_\omega\left(\boldsymbol{\phi}_\omega(G)\right)\right]. \tag{38}$$

Nonzero initialization covariance therefore produces nonzero expected variation between independent complete random graph views. This identity establishes representation diversity under random initialization but does not imply that such diversity must improve downstream affinity prediction. Its predictive utility is an empirical property of the complete reservoir–reader system.

## A6. Independent Encoder Replication versus Within-Network Widening

The $1/M$ variance law in our viewpoint arises because $M$ independently initialized complete encoders form the Monte Carlo sampling units. This construction is mathematically distinct from increasing the hidden width of one encoder.

Suppose a width-$H$ network produces coordinate-level similarity contributions $Z_1, \dots, Z_H$. The variance of their normalized sum is generally

$$\mathrm{Var}\left(\frac{1}{H}\sum_{q=1}^{H} Z_q\right) = \frac{1}{H^2}\sum_{q=1}^{H}\sum_{s=1}^{H} \mathrm{Cov}\left(Z_q, Z_s\right). \tag{39}$$

A $1/H$ variance law would therefore require additional assumptions on the covariance structure among hidden coordinates. Such independence is not implied by the present MPNN architecture because hidden channels interact through dense linear transformations, recurrent gates, normalization statistics, and repeated message propagation. Infinite-width neural networks may converge to deterministic kernel limits under suitable architectures and scaling assumptions[54], but that within-network limit is distinct from the between-network replication considered here.

A grouped implementation could be algebraically equivalent to $M$ separate encoders if all transformations remained strictly isolated across parameter blocks. For example,

$$\mathbf{W} = \mathrm{diag}(\mathbf{W}_1, \dots, \mathbf{W}_M) \tag{40}$$

would preserve the separation of $M$ independent parameter blocks when the same block structure were maintained throughout node, edge, message, recurrent, normalization, and pair transformations. Under such a construction, the Monte Carlo units remain the independent parameter blocks regardless of whether they are implemented as separate software modules or grouped into one larger computational object. A conventionally widened MPNN with cross-channel mixing does not satisfy this equivalence.

## A7. Scope of the Concentration Result

Our viewpoint characterizes the dependence of the reservoir-induced structural similarity on random encoder initialization. Specifically, the initialization variance of the scale-normalized similarity decreases as $1/M$, and the empirical similarity approaches its initialization-averaged counterpart at the stochastic rate $M^{-1/2}$.

These results do not imply that the concatenated representation itself is averaged by the supervised reader. The individual encoder blocks remain distinct coordinates in the implemented RAVEN representation. They also do not imply a monotonic decrease in downstream prediction error as $M$ increases. Increasing the reservoir size simultaneously increases the dimensionality of the supervised estimation problem while the number of labeled training complexes remains fixed. Consequently, representation coverage, estimation variance, regularization, and reader capacity may interact to produce diminishing or locally nonmonotonic predictive returns. Related multiple-random-feature models can exhibit nonmonotonic finite-sample risk behavior[33].

The theoretical result should therefore be interpreted as a representation-level concentration property with respect to encoder initialization. The predictive utility of increasing the number of frozen random graph views remains an empirical finite-sample property of the complete reservoir–reader system.